# Occupancy Network-Guided Autonomous Robotic Partial Nephrectomy

Ethan Kilmer[1†*], Pit Henrich[2†], Jiawei Ge[1†], Paul M. Scheikl[1], Laura Connolly[1], Soum D. Lokeshwar[3], Joseph Chen[1], Justin D. Opfermann[1], Kaitlyn Kumar[1], Lauren Shepard[3], Ahmed Ghazi[3], Nirmish Singla[3], Richard J. Cha[4], Kevin Cleary[4], Franziska Mathis-Ullrich[2], and Axel Krieger[1]

***Abstract*—Autonomous soft-tissue cancer surgery has been limited to interventions on organ surfaces, because current systems cannot perceive and adapt to anatomy once it deforms or is cut. We introduce the first vision-guided autonomous system capable of performing complete tumor resections for partial nephrectomy. Our system integrates conditional occupancy networks, trained entirely in a physics-based simulation, that infer full 3-D anatomy (tumor, margin tissue, and kidney) from single-view partial point clouds. These occupancy networks maintain intraoperative tracking even as tissue is cut and deformed, enabling adaptive planning and execution. The surgical platform combines a depth camera for capturing surface point clouds, dual robotic arms for electrosurgical cutting and vacuum-based tissue manipulation, and an autonomous control strategy for tumor resection. In patient-derived hydrogel phantoms under an open partial nephrectomy setting, the robot performed eight consecutive autonomous tumor resections comprising 77 electrosurgical cuts, with all cuts achieving negative surgical margins and 1.61 $\pm$ 0.48 mm mean absolute margin error. This work demonstrates, for the first time, a foundation for supervised autonomous closed-loop, imaging-driven, margin-negative tumor removal in phantoms.**

## I. INTRODUCTION

CANCER is the second leading cause of death worldwide, accounting for approximately one in six deaths (16.8%) [1]. Surgical removal, or resection, of the cancerous mass remains the primary curative treatment for many solid tumors, and involves excising the lesion with a margin of surrounding tissues (*i.e.*, the margin tissue). Surgeons aim to achieve a negative surgical margin (NSM), where pathological examination confirms no cancer cells at the specimen's margin outer surface, indicating complete tumor removal. A positive surgical margin (PSM) instead indicates residual disease and increased recurrence risk. Margin tissue sizing requires careful balance: too narrow risks incomplete resection, while too wide unnecessarily sacrifices healthy tissue and organ function.

From a surgical perspective, tumor resection requires precise cutting and careful manipulation, often using both cutting and grasping tools in coordination. With dozens of critical cuts per

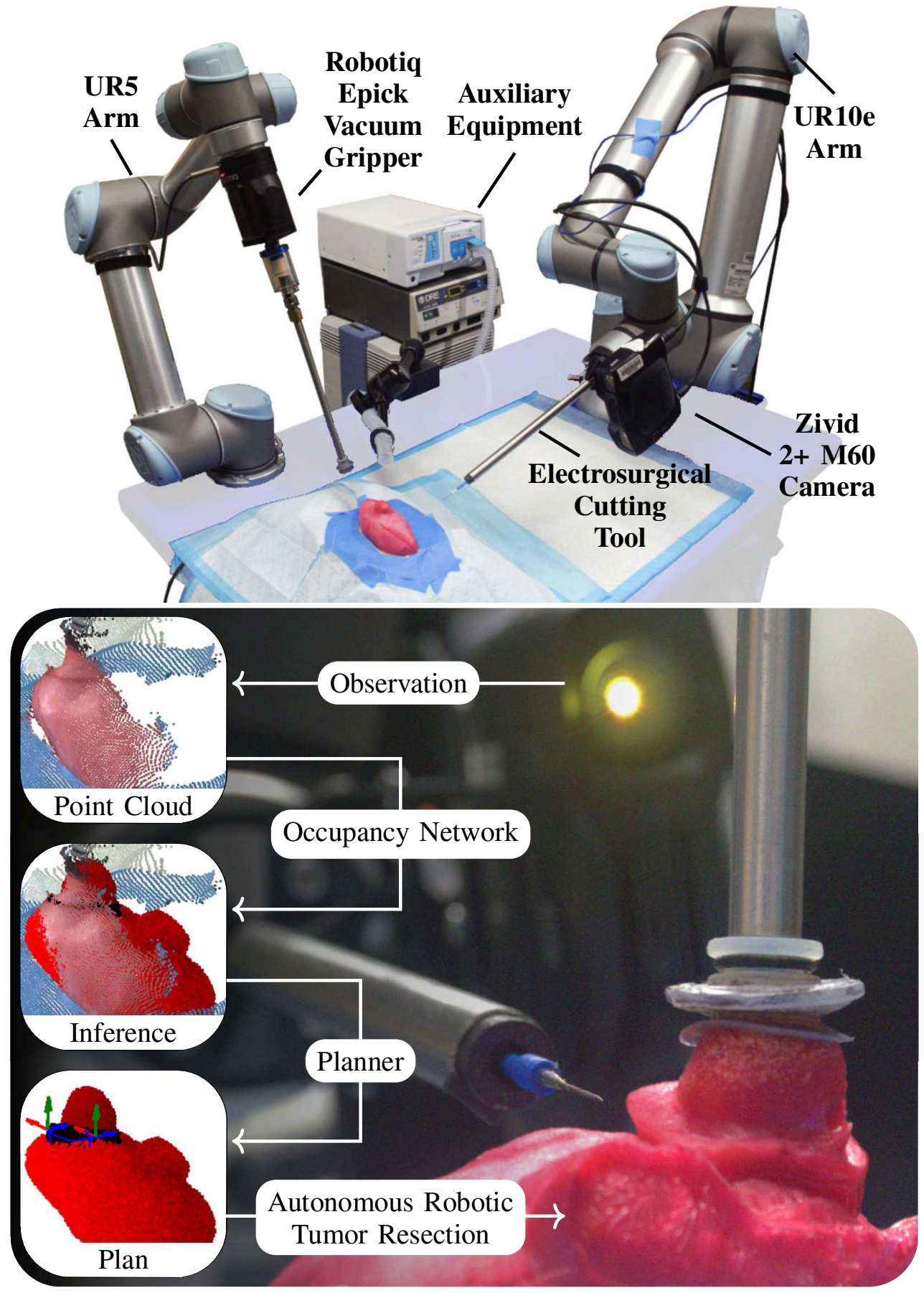

Fig. 1. Surgical robotic system for autonomous robotic partial nephrectomy. The robot's depth camera captures an observation of the surgical scene as a partial surface point cloud. A patient-specific, data-driven occupancy network provides 3-D structural information of the kidney, including information about successive stages of dissection. With the updated scene understanding, the system plans and executes autonomous robotic tumor resection on hydrogel kidney phantoms using an electrosurgical cutting tool and a vacuum gripper.

† These authors contributed equally to this work.
* Corresponding author: Ethan Kilmer.

This work was supported by the Advanced Research Projects Agency for Health (ARPA-H) under grant number AY1AX000023, and National Science Foundation (NSF/FRR CAREER) under grant number 2144348.

[1]Ethan Kilmer, Jiawei Ge, Paul M. Scheikl, Laura Connolly, Joseph Chen, Justin D. Opfermann, Kaitlyn Kumar, and Axel Krieger are with the Laboratory of Computational Sensing and Robotics (LCSR), Johns Hopkins University, Baltimore, MD 21218, USA {ekilmer1, jge9, pscheik1, lconnol8, jchen444, jopferm1, kkumar22, axel}@jh.edu

[2]Pit Henrich and Franziska Mathis-Ullrich are with the Department Artificial Intelligence in Biomedical Engineering (AIBE), Friedrich-Alexander University Erlangen-Nürnberg (FAU), 91052 Erlangen, Germany. {pit.henrich, franziska.mathis-ullrich}@fau.de

[3]Soum D. Lokeshwar, Lauren Shepard, Ahmed Ghazi, and Nirmish Singla are with the Brady Urological Institute, Johns Hopkins University, Baltimore, MD 21287, USA {slokesh1, lshepar9, aghazi1}@jh.edu, nsingla2@jhmi.edu

[4]Richard J. Cha and Kevin Cleary are with the Sheikh Zayed Institute for Pediatric Surgical Innovation, Children's National Hospital, Washington, DC 20010, USA {jcha2, kcleary}@childrensnational.org

resection, a single error causing a PSM can compromise the entire procedure. Unlike visible errors in suturing or clipping, an inadvertent cut into the tumor causing a PSM often remains undetected until pathology results return hours or days later. Achieving an NSM on the first attempt is critical, as PSMs necessitate complex salvage strategies and worsen oncologic outcomes. Direct tumor grasping is avoided due to the risk of rupture and spread; instead, surgeons retract adjacent healthy tissue to minimize stress and preserve oncologic safety.

From an imaging and guidance standpoint, tumor resections typically rely on preoperative computed tomography (CT) or magnetic resonance imaging (MRI) images. Surgeons mentally align these images with intraoperative visual and tactile cues, sometimes aided by ultrasound (US) or near-infrared fluorescence (NIRF) imaging [2]. As surgery progresses, this mental mapping becomes unreliable. Tissue deformations, and cutting continuously alter the anatomy, making it inconsistent with preoperative images. The mental mapping becomes especially unreliable during deep margin dissection, where the tumor is completely occluded by surrounding margin tissue and no longer visible, creating a blind zone, as illustrated in Figure 3. NIRF signals penetrate only a few millimeters of soft tissue and vary with dye uptake, excitation, and sensor sensitivity, so apparent signals cannot be translated into a reliable subsurface distance for precise cut-plane guidance [2]. US offers stronger subsurface resolution and is therefore preferred for deep margin assessment in several resections, such as intraoperative US during breast-conserving surgery [3].

### A. Autonomous Robotic Tumor Resection

Robot-assisted surgery (RAS) has gained widespread adoption since the early 2000s across diverse procedures, including tumor resections. Through teleoperation, it improves ergonomics, visualization, precision, and dexterity for surgeons. However, surgical procedures remain manually performed, with outcomes still heavily dependent on individual surgeon skill, experience, and performance. While teleoperation reduces surgeon fatigue and burnout, it cannot fully eliminate these factors, which contribute to major medical errors and compromise surgical outcomes [4].

Integrating autonomy into surgery has emerged as a promising research direction aimed at delivering consistently precise outcomes. Most efforts have focused on structured interventions such as orthopedic, refractive, percutaneous, and stereotactic radiosurgeries [5]. More recently, autonomy has been extended to select subtasks in soft tissue surgery, such as suturing in anastomosis and vessel clipping or cutting in cholecystectomy, using either model-based approaches that prioritize predictability and safety, or learning-based methods that improve adaptability and scalability [6], [7]. Unlike RAS, autonomous robotic surgery faces additional translational barriers, particularly ethical and regulatory concerns, and remains largely confined to academic research.

Tumor resection remains a difficult and largely unexplored field in autonomous soft tissue surgery. Most current research relies on color, US, or NIRF imaging, but only achieves initial surface incisions [8]–[13]. Only three studies have achieved autonomous removal of 3-D tumor-like masses [14]–[16]. However, these three approaches bypass the most challenging deep margin tissue dissection phase. McKinley et al. [14] remove a rubber tumor from a soft pad by simple peeling (*i.e.*, targeting a tumor without a margin). Ge et al. [15] treat tumors as zero-thickness shapes on porcine tongue surfaces, cutting uniform extruded volumes after assigning lateral and depth margin (*i.e.*, targeting a margin without a tumor). Smith et al. [16] perform palliative central airway obstruction resections to restore airway patency, where incomplete tumor removal is clinically acceptable (*i.e.*, targeting symptom relief without requiring an NSM).

### B. Kidney Cancer and Partial Nephrectomy

Tumor resection techniques vary by surgical specialty, cancer type, stage, spread, and anatomical location. This work centers on kidney cancer and its surgical treatment.

Kidney cancer ranks as the 14th most common cancer globally, with an estimated 434 419 new cases and 155 702 deaths in 2022 [1]. Renal cell carcinoma (RCC) is the predominant histological subtype, accounting for over 90% of kidney cancer cases [17]. Diagnosis involves blood or urine tests, biopsies, or medical imaging such as US, CT, or MRI. Treatment for advanced RCC (stages 2-4) often includes radical nephrectomy (RN), removing the entire kidney and surrounding tissue, followed by targeted therapy or immunotherapy for metastatic disease. In contrast, early-stage RCC (stage 1, typically $<$ $70\,\mathrm{mm}$ in diameter) is primarily treated with partial nephrectomy (PN), which removes only the tumor and some surrounding tissue. PN, when compared to RN, better preserves renal function while maintaining equivalent cancer control [18]. Achieving an NSM is a strong prognostic factor for improved postoperative function and reduced recurrence risk [19].

Three surgical approaches to PN exist for stage 1 RCC: open (OPN), laparoscopic (LPN), and robot-assisted (RAPN). While all three show statistically similar NSM rates, OPN offers shorter operative and warm ischemia (duration without blood flow) times, whereas LPN and RAPN provide shorter hospital stays and fewer complications, with RAPN additionally reducing blood loss compared to LPN [20]. However, clinical outcomes remain suboptimal, with reported PSM rates of $5.5\,\%$ and overall complication rates of $18.7\,\%$ [21].

While tumor resection is the critical curative step, it represents only one component of the complete PN procedure: preoperative imaging, surgical access (large incision for OPN or keyhole incisions with pneumoperitoneum for LPN/RAPN), fat dissection for kidney mobilization, vascular clamping, tumor resection, tissue fragment removal, kidney repair, blood flow restoration, fat and skin closure.

Intraoperative US and NIRF are commonly used immediately before tumor resection to localize the lesion and mark the surface margin [2], [22], [23]. NIRF may also be assessed after resection for residual tumor or suspected PSMs. Neither modality is typically reacquired during deep margin dissection. Reacquiring US under warm ischemia requires instrument exchanges that prolong operative time and disrupt workflow; moreover, cuts create air gaps through which US waves cannot propagate, preventing imaging across the resection field.

Surgeons therefore scan once for surface marking, then rely on visualization and mental mapping. Continuous subsurface guidance during this phase remains unavailable in PN.

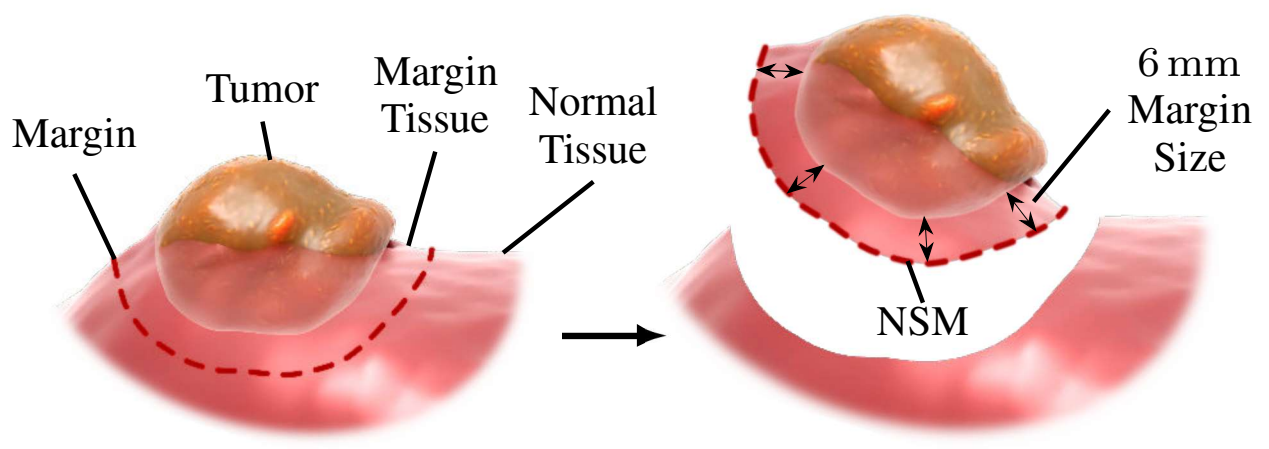

Fig. 2. Illustrative image of the tumor resection task in this study: achieving a negative surgical margin (NSM) with a precise 6 mm margin size.

### C. Task Description

This study proposes a *Multi-shot* occupancy-guided autonomous robotic system for OPN under human supervision. The term *Multi-shot* refers to the repeated intraoperative scene capture and 3-D anatomy inference throughout a resection. An illustration of our resection task is shown in Figure 2. Most RCC cases involve partially exophytic tumors that protrude from the kidney surface [24]. We target the OPN setting, in which the kidney is surgically mobilized and surrounding organs are retracted out of the tool workspace [23], [25], [26]. This choice avoids minimally invasive surgery (MIS) constraints, such as limited workspace and surrounding-organ management, allowing us to focus on the core challenge of subsurface guidance. All experiments in this paper follow this OPN setting. The primary goal is to achieve an NSM by aiming for a 6 mm margin tissue size, a clinically appropriate choice based on surgical guidelines [27], [28].

### D. Contributions

First, we introduce a Multi-shot occupancy-guided intraoperative inference workflow for tumor resection, which we validate in an OPN setting. This is the first approach able to track a resection throughout the entire phantom tumor removal process, from surface incision through deep margin dissection, including stages where the tumor itself becomes occluded by margin tissue as illustrated in Figure 3. Trained entirely in physics-based simulation, the workflow repeatedly infers complete 3-D anatomy (tumor, margin tissue, healthy tissue) from single-view partial point clouds despite deformation and cutting, and supports online re-planning throughout the procedure. We validate the occupancy-guidance accuracy by comparing predicted 3-D anatomy to reference CT volume, achieving $2.13 \pm 1.04\,\mathrm{mm}$ tumor centroid error across 16 diverse resection states.

Second, we integrate this perception workflow in a hybrid learning and model-based autonomy stack optimized for tumor resection surgeries. We combine learning-based patient-specific adaptive intraoperative imaging, model-based state machines for predictable workflow control, and human supervision checkpoints for each dynamically updated dissection plan. This design specifically addresses tumor resection's unique constraint: cutting errors causing a PSM often cannot be intraoperatively detected (as detection may require pathological assessment under microscopy after complete tumor removal) or corrected (because microscopic residual cancer cells, spillage, and dissemination may have already occurred and are irreversible). With dozens of high-stakes cuts per resection, a single error can compromise the entire procedure, making first-attempt success critical.

Third, we achieve the first precise autonomous robotic OPN (AROPN). Using tissue-mimicking hydrogel phantoms derived from anonymized patient CT data, we performed eight consecutive autonomous tumor resections comprising 77 electrosurgical cuts; all cuts succeeded, achieving NSMs in all eight procedures (100 % success rate). The mean absolute margin error was $1.61 \pm 0.48\,\mathrm{mm}$, and the mean signed margin error (bias) was $+0.41 \pm 1.04\,\mathrm{mm}$ (positive indicates larger-than-planned clearance).

## II. Related Work

### A. Autonomous Tumor Resection

While autonomous surgery began in 1991 with the PROBOT robot for transurethral resection of the prostate (TURP) procedures [29], autonomous tumor resection in soft tissue started much later. McKinley et al. in 2016 first demonstrated tumor mass removal using the da Vinci Research Kit (dVRK) to palpate, expose, and remove rubber tumors from silicone pads [14]. Subsequent work has primarily focused on surface incisions with various imaging modalities: Saeidi et al. and Ge et al. in 2019 used Smart Tissue Autonomous Robot (STAR) with NIRF guidance on porcine tongues [8], [9]; Marahrens et al. in 2024 used dVRK with US guidance on porcine livers [11]; Kilmer et al. in 2025 used Autonomous System for Tumor Resection (ASTR) with NIRF guidance on kidney phantoms [12]; Kam et al. in 2025 used a KUKA manipulator with RGB-D and CoTracker-based tracking for dynamic cutting speed adjustment on porcine tongues [13]. As discussed in Section I-A, only three studies have achieved complete tumor mass removal [14]–[16], none of which address intraoperative imaging under the tissue deformation, cutting, and occlusion that occur during deep margin dissection. This gap motivates our focus on maintaining resection tracking throughout the entire tumor removal procedure.

### B. Soft Tissue Registration

In our context, deformable registration seeks to align a preoperative 3-D model with an *in-situ* point cloud observation even when the target changes shape.

Under the assumption that deformations are small, a rigid registration approach can be used. Iterative Closest Point (ICP) [30] iteratively estimates a transformation, through a loss minimization, which aligns the preoperative model with the point cloud observation. But this process is prone to local minima and fails with increasing levels of deformation.

Coherent Point Drift (CPD) [31] allows the preoperative model to deform during registration to match the point cloud observation. While CPD can align point sets under deformations, poor initialization leads to suboptimal local minima.

Data-driven methods [32], [33] learn a dense deformation field that morphs a preoperative 3-D model so its surface matches the observed point cloud. Although these networks embed biomechanical priors, they still depend on a rigid pre-alignment. All previously mentioned methods assume a solid rigid alignment. However, rigidly aligning a rigid prior to a deformable object is inherently ill-posed.

Recent work re-frames the registration task: Rather than explicitly optimizing a transformation or warp field, a neural occupancy network is conditioned directly on the observed point cloud to produce a fully deformed 3-D object in a single forward pass. This eliminates the need for an initial rigid alignment. The output replaces the prior object and can be used directly for downstream tasks.

Early approaches demonstrated the potential of such methods [34], [35]. These neural occupancy network-based methods are trained object specific and fully supervised. During training, they learn to associate single-view surface observations with the full 3-D shape of an object. Therefore, they learn to represent a 3-D object that explains the single-view surface observation. However, during training, these approaches do not use a physics-based simulation to generate the training examples. This results in a limited real-world accuracy. LUDO [36] addresses this limitation by using a finite element method (FEM)-based simulation framework to generate training data. In robotic experiments across three deformable objects, LUDO demonstrates that neural occupancy networks can guide a needle to autonomously puncture internal regions of interest (*e.g.*, percutaneous biopsy) with a high success rate. We want to briefly clarify that data-driven methods such as [34]–[36] estimate the spatial shape of known 3-D objects. This task differs from scene reconstruction, as performed by NeRF [37] for example, that estimate and reconstruct the shape of unknown objects from observations.

### C. Tumor Model

Appropriate tumor models are essential yet difficult to establish for surgical oncology research. While actual patient tumors offer the highest fidelity, research involving human subjects is limited to individuals already diagnosed with cancer and tightly regulated under ethical oversight. Cadaver specimens with tumors are too rare to be considered for many studies. Consequently, *in vivo* animal models, particularly rodents implanted with human cancer cell lines, primarily serve biological studies, but their small size limits surgical applicability. Larger animals like pigs enable more realistic surgical procedures through two approaches: genetic tumor models and artificial mimics. Schachtschneider et al. [38] developed the Oncopig cancer model, a transgenic swine platform enabling site-specific tumor induction for translational research. For artificial mimics, Kawai et al. [39] injected muscle paste into live pig lungs to simulate CT-visible tumors for radiofrequency ablation training, while Taylor et al. [40] developed agarose-based renal tumor mimics for both *in vivo* and *ex vivo* use with US visibility.

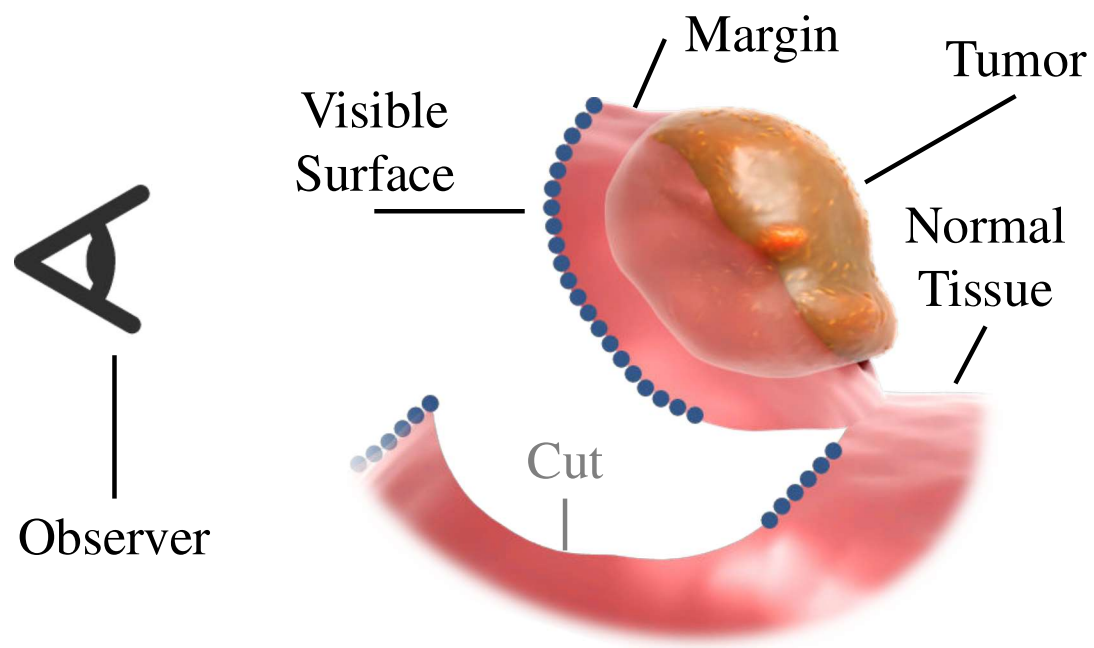

Fig. 3. The tumor surface becomes occluded by margin tissue. To access the tissue under the tumor, the tumor needs to be pulled away from the resection. As a result of this, the tumor itself may be occluded from the observers view.

Synthetic phantom models further support surgical training and evaluation. Melnyk et al. [41] used hydrogel casting based on patient CT scans to fabricate perfused kidney phantoms with embedded tumor analogs. Similarly, Monda et al. [42] created silicone-based renal tumor phantoms using 3-D printed molds for robotic nephrectomy training. High-quality synthetic phantom models are well-suited for research due to their reproducibility (*i.e.*, the same tumor phantom can be produced multiple times). Further, they allow for surgical planning and robotic training without ethical constraints.

## III. System Overview and Task Formulation

The central technical contribution of this work is a Multi-shot occupancy-guided imaging workflow for autonomous tumor resection, validated in an OPN setting. During the procedure, the system repeatedly captures the exposed kidney surface, segments the target anatomy, infers the full 3-D anatomy with a patient-specific occupancy model. This model is used to generate an updated dissection plan, and executes that plan with a dual-arm robotic platform under human supervision. The occupancy-guided inference provides the adaptive intraoperative scene understanding, whereas the robot system provides the means to carry out the cuts and tissue manipulation.

### A. Multi-shot Task Decomposition

Recognizing intervention and imaging challenges during tumor resection as introduced in Section I, the Multi-shot workflow decomposes the procedure into two phases. The first is circumferential surface incision (CSI), the initial cuts around the tumor; the second is progressive deep dissection (PDD), the later cuts beneath the tumor. In CSI phase, the exophytic tumor surface remains directly visible, allowing the preoperative model to be aligned with this surface for accurate guidance. The cutting tool then performs a planned surface incision along a mapped contour. In PDD phase, the tumor becomes progressively occluded by margin tissue until it may be fully out of view, as illustrated in Figure 3. The occlusion, combined with tissue deformations and morphology-altering cuts, prevent the preoperative model from being fitted directly to the intraoperative scene. In PDD phase, grasping and cutting tools must work in coordination through iterative retraction and cutting to achieve complete tumor removal.

### B. Anatomical Representation for Guidance

The Multi-shot workflow represents the resection scene through four patient-specific anatomical structures: the kidney ($K$), the circumferential cut ($C$), the margin tissue ($M$), and the tumor ($T$). Each structure serves a distinct guidance role. $K$ provides the surrounding organ geometry for anatomical context, $C$ defines the target region for the initial CSI, $M$ defines the intended resection shell around the tumor, and $T$ identifies the tumor location and therefore the boundary of oncologically unsafe cuts. During CSI, the planner operates on $(K, C, M, T)$, while during PDD the circumferential cut is assumed to be removed and the guidance problem reduces to $(K, M, T)$ under progressive deformation and occlusion. These structures are derived from preoperative CT.

## IV. Multi-shot Occupancy-Guided Intraoperative Inference

### A. Intraoperative Imaging and Guidance

The Multi-shot workflow requires an intraoperative inference module that localizes the tumor and provides updated anatomy for planning each cut. A robotic system can contribute partial surface point clouds of the kidney through its depth sensor. However, this data lacks the full 3-D structural information needed to safely guide resection of partially endophytic tumors. The preoperative patient CT scan provides the 3-D information, however, registering the scan to the point cloud, particularly after extensive cutting and deforming has altered the kidney, remains a challenge. Current deformable object registration methods have no way to handle cuttable objects. We therefore introduce a cut-aware framework based on conditional occupancy networks for obtaining 3-D structural information. This framework enables prediction of both the intact organ structure and its progressive states during surgical cutting. As an instantiation, we use the state-of-the-art LUDO [36], which generates a 3-D object representation conditioned on an observation rather than by deformable registration. We show how such conditioned occupancy networks can be trained to reason about cuttable objects, provided that cutting behavior is learned during training.

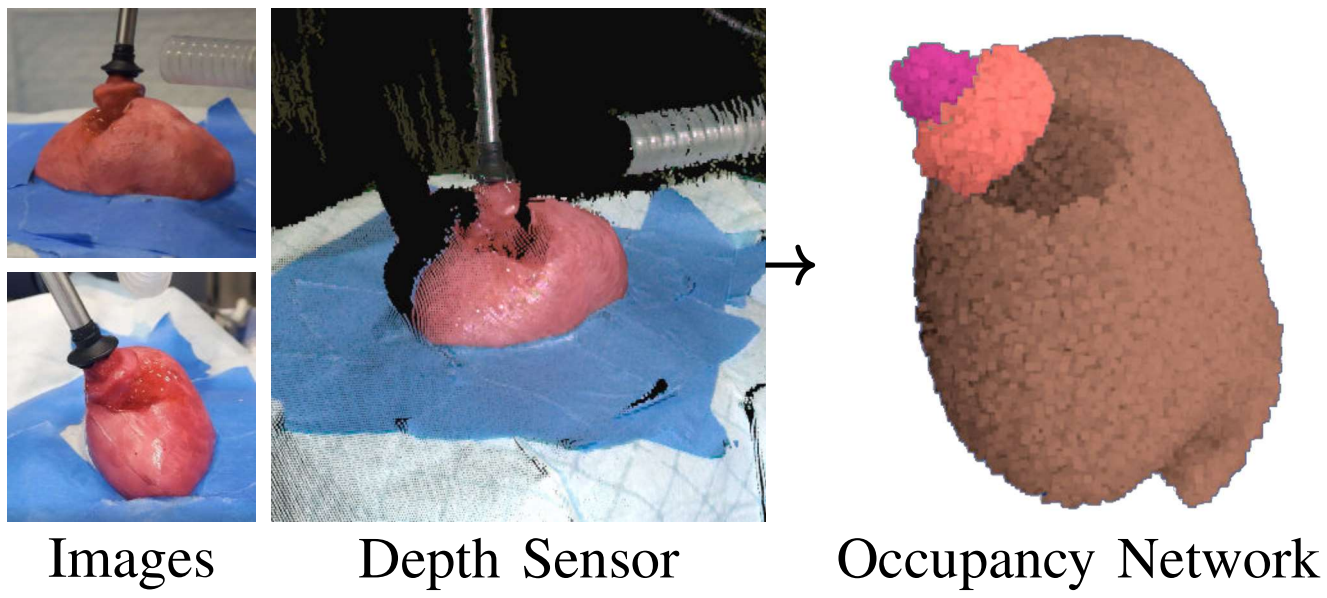

Fig. 4. Occupancy network output used for scene understanding. The left shows images of a kidney after 10 progressive depth cuts. The middle shows the depth sensor observation. The right shows the output of our occupancy network.

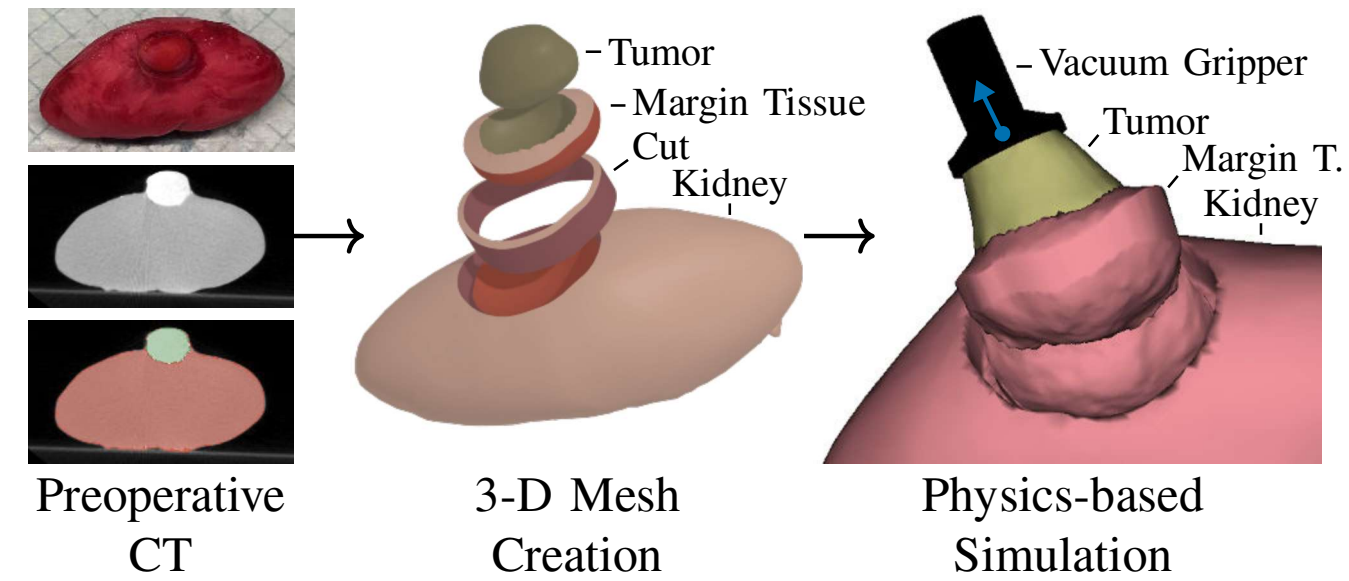

Fig. 5. Intraoperative tumor resection states are generated in physics-based simulation. A preoperative CT of the patient (hydrogel kidney phantom) is obtained and segmented. The segmentation is used to create 3-D meshes of the tumor, margin tissue, circumferential cut, and kidney. These meshes are imported into a physics-based simulation with a cut modeling the tumor resection. The vacuum gripper retracts the tumor following the expected real world trajectory (blue arrow).

In our context, a conditional occupancy network learns a 3-D representation of an object, conditioned on an observation[1]. In this application, the observation is a surface point cloud of the kidney that is being operated on. The network adapts its internal 3-D representation based on what it observes in the point cloud. For example, if the surface point cloud contains a cut, the resulting 3-D representation will reflect a kidney with a corresponding cut, examples are shown in Figure 1 and Figure 4.

We train two conditional occupancy networks, one per resection phase. The CSI model provides information about where to perform the initial surface incision, whereas the PDD model provides information for the guidance of the subsequent deep dissection cuts. Note, the CSI phase could also be guided using a traditional registration approach, as no deformation is assumed for the kidney. However, conditioned occupancy networks have been shown to achieve state-of-the-art performance for anatomical structure tracking [36], and using a unified perception representation across both phases provides a consistent framework for surgical guidance. The data pipeline for training involves creating patient-specific 3-D meshes from preoperative scans, simulating the tumor resection, and capturing surface point cloud and occupancy ground-truth data pairs for training.

### B. Occupancy Network: 3-D Mesh Creation

To train a conditioned occupancy network, we must have a 3-D representation of the scene that is to be learned. We obtain the 3-D kidney object from a CT scan of a patient. After segmentation, the kidney and tumor segmentation masks are converted into surface meshes. To geometrically instantiate the guidance structures introduced in Section III, we construct the four components $(K, C, M, T)$, as visualized in Figure 5.

These structures are created through a sequence of Boolean operations. First, we place a hollow cylinder concentrically around the tumor, with inner and outer walls offset by $5.5\,\mathrm{mm}$ and $7.7\,\mathrm{mm}$, extending $10\,\mathrm{mm}$ into the kidney. The intersection of this cylinder with the kidney defines the circumferential cut ($C$). Subtracting the cylinder produces a kidney-with-cutout mesh. Next, we enlarge the tumor by $6\,\mathrm{mm}$ along vertex normals and intersect it with this kidney-with-cutout mesh to obtain the margin tissue ($M$). To note, the selection of $6\,\mathrm{mm}$ margin tissue size is clinically informed and introduced in Section I-C. Subtracting the margin tissue yields the final kidney mesh ($K$). Finally, we remove degenerate triangles introduced by the Boolean operations while preserving vertex consistency at mesh boundaries. The cleaned meshes are then converted into tetrahedral meshes using Gmsh.

[1] You can imagine that the model watches (single-view surface point cloud) objects deforming or being cut, and learns to associate the observation with the 3-D state of the object.

### C. Occupancy Network: Physics-based Simulation

While the composition of the meshes $(K, C, M, T)$ allows us to train an occupancy network to understand the scene for the CSI phase, we need to train a separate PDD model. Gathering data that reflects the PDD phase requires simulating cutting and deforming meshes $(K, M, T)$, following progressive iterations of deep dissection. Note, $C$ is removed since this portion has already been cut away.

Cutting volumes in virtual simulation is an active area of research. Existing methods include tetrahedral element removal, mesh refinement, and particle-based approaches [43]–[45]. Since these methods are dependent on mesh resolution and particle density, and have limited previous demonstrations carving complex contours such as those seen in tumor resection, we achieve cutting through a different approach. Note, we want to obtain 3-D meshes that represent states after cutting actions. The cutting actions themselves do not need to be simulated continuously, only the state after a cut was performed needs to be obtained.

Figure 1 and Figure 9.h-i show the robot performing a PDD cut beneath the margin tissue.We develop a simulator that takes input meshes $(K, M, T)$ and produces a kidney with these cuts. A virtual cutting plane, perpendicular to the cutting tool approach vector, is positioned according to a desired resection percentage ($0\%$ meaning no cuts have occurred and $100\%$ being complete tumor removal). Shared vertices and faces of $K$ and $M$ beyond the plane (away from the cutting tool) are merged, leaving the rest of the margin tissue cut. Random $\pm 5^\circ$ rotations are applied to the plane to account for real world cutting path variability. The tumor $T$ is attached to margin tissue $M$ by merging shared vertices and faces.

The occupancy network should learn to handle deviations from the intended cutting surface caused by resection errors. Therefore, we propose a surface randomization of the cut surface as an augmentation. For this augmentation, we apply layered Perlin noise[2] to the cut surface vertices (displacement along the CSI angle) to generate realistic variations, as shown in Figure 6. One low-frequency function adds large-scale shape variations (up to $3\,\mathrm{mm}$), while a high-frequency function introduces finer detail with a randomized amplitude between $0$ and $1\,\mathrm{mm}$. Note that noise is only applied to the cut points, the uncut points remain unchanged.

Next we use a physics-based simulation of the tumor resection scene using the FEM physics engine Simulation Open Framework Architecture (SOFA) [46]. SOFA is an open-source framework for physics-based simulations with an emphasis on biomechanics and robotics. In addition to our experience with SOFA, the framework is also widely applied to the simulation of soft robotic actuators, which depend on accurate modeling of internal deformations [47]–[49]. Modern FEM-based simulators such as SOFA exhibit a sufficiently small sim-to-real gap to support the transfer of deformable object manipulation strategies from simulation to real-world settings [50] and to train accurate data-driven methods for tracking tumors embedded inside of objects [36].

[2]Perlin noise can be thought of as smoothed TV static.

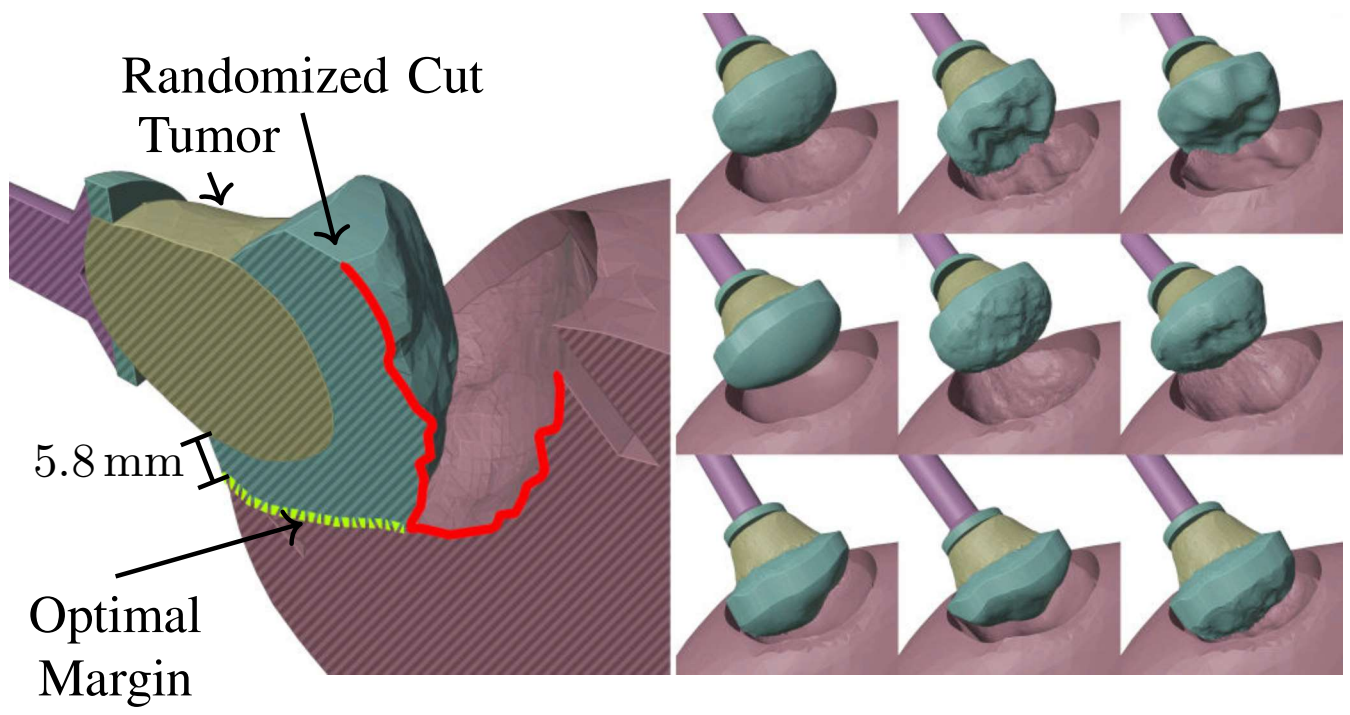

Fig. 6. Training data induces guidance towards the optimal cutting path. On the left, a cross-sectional view of the kidney illustrating the ongoing cut. The already cut surfaces are randomized, indicated by the red lines. Regions that remain to be cut are not randomized. The green dashed line marks the optimal margin used for guidance. This optimal margin is approximately $5.8\,\mathrm{mm}$ thick (below the $6\,\mathrm{mm}$ target due to tissue compression). This strategy exposes the occupancy network to variations in the cut surface, with the expectation of making it robust to irregular cuts while always providing structural information about the optimal margin. On the right, several examples show how the cut surface is randomized.

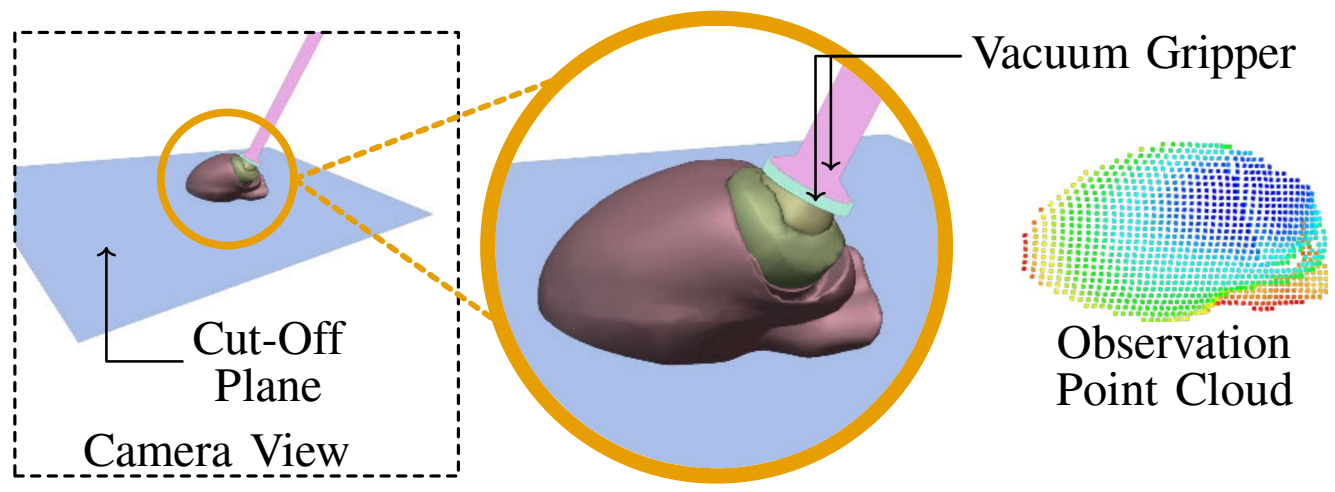

Fig. 7. Virtual scene used to generate data for the occupancy network. A deformed and cut kidney instance is loaded. The *Cut-Off Plane* and the *Vacuum Gripper* are used as masks to occlude parts of the kidney. The kidney is rotated randomly and the Cut-Off Plane is moved randomly along its normal. The resulting *Observation Point Cloud* seen by occupancy network during training is shown on the right.

After cutting, we use the mesh to generate a tetrahedral co-rotational FEM force field with material parameters offered by Melnyk et al [41]. Figure 5 shows the robot's vacuum gripper as a surface mesh with a constraint to the tumor top to simulate a grasp. The vacuum gripper lifts the tumor following Algorithm 1, further explained in Section V. The retraction incrementally pauses to save the vacuum gripper and ($K$, $M$, $T$) as surface meshes. $10\,072$ triplets of $(K, M, T)$ are generated to train the PDD model for each patient.

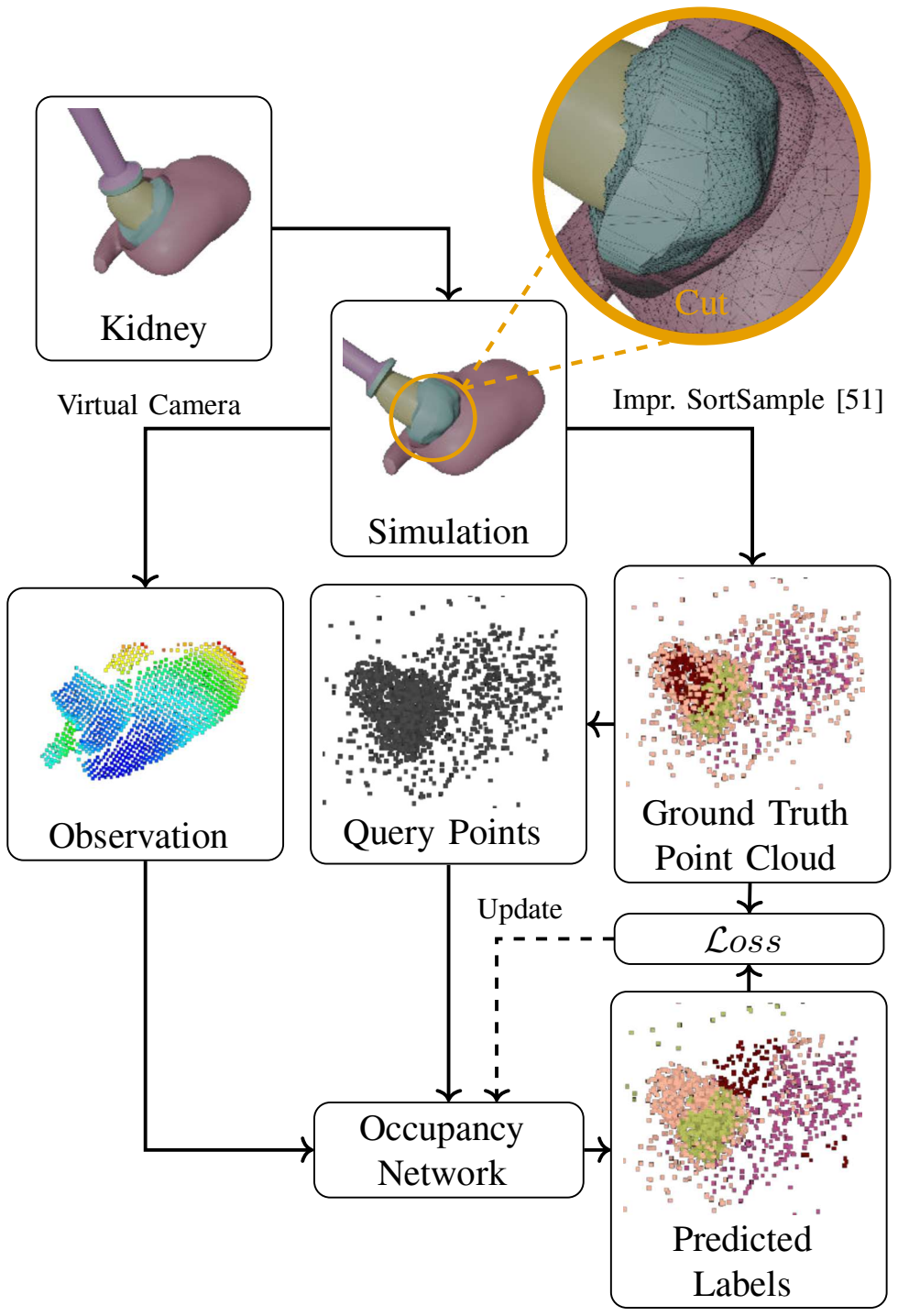

Fig. 8. PDD phase training loop. The patient-specific *Kidney* is deformed and cut in *Simulation*. A virtual camera generates an *Observation*. Simultaneously, a labeled *Ground Truth Point Cloud* is obtained using the *improved SortSample* [51], where each point is assigned to the segment it occupies. The unlabeled *Query Points* are obtained from the *Ground Truth Point Cloud* by removing the labels. The *Occupancy Network* is conditioned on the *Observation* and predicts labels for query points, resulting in the *Predicted Labels*. A *Loss* (e.g., cross-entropy) is computed between the *Predicted Labels* and *Ground Truth Point Cloud* and used to update the *Occupancy Network*.

### D. Occupancy Network: Training Models

As a final step in data generation, we use the meshes obtained from the simulation to produce surface point clouds. These point clouds resemble what the robot may observe during the CSI and PDD phases and are paired with corresponding ground truth occupancy point clouds. To produce this data, we use a virtual environment consisting of a camera, a vacuum gripper, and a cut-off plane. Depending on whether we train the CSI or PDD model, we add either the meshes $(K, C, M, T)$ or $(K, M, T)$ to the scene.

For the PDD model, a randomly cut and deformed mesh from the SOFA simulation is loaded (one of $10\,072$). The cut-off plane and the vacuum gripper are used to mask out (only from the observation) parts of the kidney, teaching the network that parts of the anatomy may be occluded during surgery. The kidney orientation is randomized by up to $\pm 35°$ along each axis, ensuring robustness to different poses. The cut-off plane is placed at the center of the kidney and shifted randomly by up to $\pm 25\,\mathrm{mm}$ along its normal. An example of the virtual scene is shown in Figure 7. The camera captures depth images with a resolution of $256 \times 256$. We use Improved Sort Sample (ISS) [51] to obtain a ground truth occupancy point cloud.

The datasets for the CSI and PDD phases contain $32\,000$ and $50\,000$ training examples respectively, each consisting of an observation and a corresponding occupancy point cloud.

TABLE I
TRAINING CONFIGURATION AND COST FOR THE OCCUPANCY MODELS. PARAMETERS ARE SHARED ACROSS PATIENT MODELS WITHIN EACH PHASE. 'DEFS.' DENOTES THE NUMBER OF SIMULATED DEFORMATIONS, 'B. SZ.' THE BATCH SIZE, 'GEN. TIME' THE DATA-GENERATION TIME, AND 'TRAIN T.' THE MODEL TRAINING TIME. HARDWARE: AMD RYZEN 9 7950X (CA, USA), NVIDIA GEFORCE RTX 4090 (CA, USA)

| Phase | Samples | Defs. | Epochs | B. Sz. | Gen. Time | Train T. |
|---|---|---|---|---|---|---|
| CSI | $32\,000$ | 0 | 500 | 32 | $0.5 \pm 0.1\,\mathrm{h}$ | $5 \pm 0.5\,\mathrm{h}$ |
| PDD | $50\,000$ | $10\,072$ | 500 | 32 | $1.7 \pm 0.4\,\mathrm{h}$ | $7.8 \pm 1\,\mathrm{h}$ |

These simulated observations reproduce the appearance and occlusions of real surgical depth camera point clouds, ensuring that the trained models can transfer to the robotic setup. We then train two conditional occupancy networks instantiated with LUDO [36], using the reference hyperparameters. Since the training configuration is identical across patient models, we report shared parameters for each phase together with the corresponding training cost. The total number of training samples and the number of simulated deformations used to generate the PDD training data are summarized in Table I. An overview of how we train our occupancy networks is shown in Figure 8.

## V. Autonomous Resection Workflow

Once trained, the occupancy networks are integrated into our autonomous control system. We use a FSM-based planner. This planner consists of modular autonomous states, which correspond to phases in a surgical workflow. By providing transition conditions between these states, we obtain an FSM-planner that can achieve autonomy, as shown in Figure 9. Human supervisors oversee state transitions, for example between cutting plan generation and execution, with the option to trigger re-planning to prevent errors that might cause a PSM.

We deliberately avoid fully end-to-end learning-based approaches because of their lack of next-state safety guarantees. Approaches such as SRT-H [7] and Yell At Your Robot [52] demonstrate impressive adaptability but rely on error detection and corrective behaviors. In tumor resection, however, errors such as PSMs are frequently unobservable intraoperatively and may be oncologically irreversible once committed [19]. Our FSM-planner, integrating the occupancy network, enables pre-execution intervention to prevent unsafe plans.

The FSM-planner is implemented in ROS 2 using a multi-threaded executor running at a 10 Hz loop rate to maintain responsiveness during long-running operations. Each state triggers asynchronous routines such as motion planning or point cloud acquisition, with progress monitored through topic feedback and callback flags.

State 1 initializes the dual-arm system by recording home positions for post-procedure return (Figure 9.a), loading the predefined scene capture pose as well as wrist-up and wrist-down joint values of the electrosurgical arm (Figure 9.d-e). States 2-4 handle scene capture: positioning the camera (State 2), capturing a point cloud of the exposed kidney surface (State 3, Figure 9.b), and segmenting kidney points using Segment Anything Model 2 (SAM 2) [53]. During the first capture, the human supervisor clicks a pixel on the kidney in the 2-D image; this coordinate is saved and reused as the SAM 2 point prompt for all subsequent captures. The resulting segmentation mask is eroded using a 5×5 pixel kernel to mitigate RGB-D edge artifacts (see Figure 17 in the Appendix), and the corresponding 3-D points are extracted. The segmented point cloud is then processed through the trained occupancy network model to generate a labeled dense point cloud distinguishing points $(K, C, M, T)$ for CSI phase or points $(K, M, T)$ for PDD phase (State 4, Figure 9.c). Here, *occupancy inference* denotes the occupancy-network inference and anatomy labeling stage only, whereas the *end-to-end* runtime denotes the full sequence from point cloud capture through labeled anatomy output used by the planner.

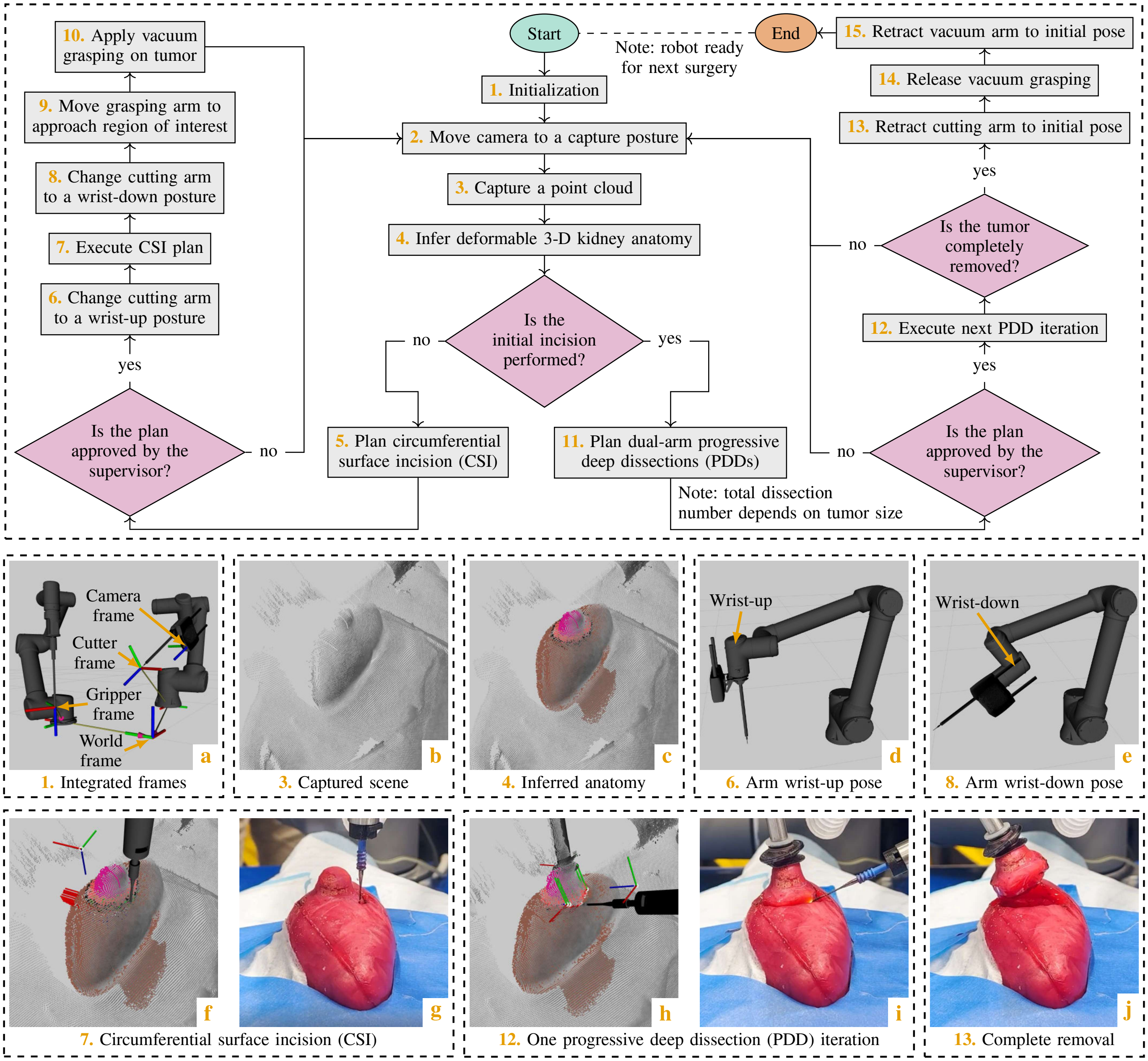

Fig. 9. 15-state finite-state machine (FSM) state diagram of the dual-arm autonomous tumor resection workflow for open partial nephrectomy (OPN). Top: FSM state diagram showing the sequence from initialization through circumferential surface incision (CSI, States 5–7) and progressive deep dissection (PDD, States 11–12) to tumor removal. Supervisor approval gates are shown before CSI and each PDD iteration. Bottom: Representative images corresponding to key states. (a) Coordinate frames for camera, cutter, gripper, and world. (b) Captured colorless point cloud scene. (c) Inferred 3-D anatomy with tumor (pink), margin tissue (orange), and kidney (tan). (d–e) Cutting arm wrist-up and wrist-down poses. (f–g) CSI execution showing planned trajectory and electrosurgical cutting. (h–i) One PDD iteration with vacuum retraction and deep margin dissection. (j) Complete tumor removal with specimen held by vacuum gripper.

State 5 generates the CSI path by analyzing incision rim points $(C)$ to identify the deepest penetration contour, computing the cutting tool orientation as 97% gravity-aligned and 3% directed toward the contour centroid, with the path visualized in RViz2 for supervisor approval, while rejection triggers return to State 2 for re-planning.

States 6-8 execute the approved incision: transitioning to wrist-up posture for near-vertical cutting (States 6, Figure 9.d),

executing the CSI (States 7, Figure 9.f-g), then returning to wrist-down posture (States 8, Figure 9.e). States 9-10 apply vacuum grasping by positioning the gripper $20\,\mathrm{mm}$ above the tumor points apex ($T$), descending $20\,\mathrm{mm}$ plus bellows height for firm contact, activating vacuum, then lifting $5\,\mathrm{mm}$ plus half the bellows height while tilting $10^\circ$ backward and moving $5\,\mathrm{mm}$ laterally away from the cutting arm to establish initial tissue tension.

Following incision completion, the system returns to State 2 to capture updated geometry before State 11 plans PDDs by partitioning the margin tissue points ($M$) into $3\,\mathrm{mm}$-spaced parallel slices perpendicular to the dissection direction, computing cutting paths via 2-D convex hull analysis of projected margin tissue points (Figure 9.h), and preparing progressive vacuum retraction that gradually pulls the tumor away from the kidney bed between dissection iterations. State 12 executes each PDD iteration through coordinated dual-arm motion: the cutting arm follows the planned dissection path with its tool orientation progressively transitioning from $25^\circ$ to $15^\circ$ downward tilt, while the vacuum arm incrementally retracts following Algorithm 1, which implements a square-root profile in the vacuum arm's base frame that smoothly transitions lateral movement from $3\,\mathrm{mm}$ to zero, vertical lift from zero to $1\,\mathrm{mm}$, and backward tilt from $5^\circ$ to zero across all iterations, ensuring adequate tissue tension throughout the dissection (Figure 9.h-i), after which the system returns to State 2 for planning next PDD iteration based on updated anatomy. The dissection loop continues for a total number of iterations equal to the tumor width divided by $3\,\mathrm{mm}$ slice thickness, terminating when all iterations complete or upon supervisor confirmation of complete tumor removal (Figure 9.j).

**Algorithm 1** Progressive Vacuum Retraction Profile

**Require:** Current dissection iteration $i \in \{1, 2, ..., n\}$, total iterations $n$
**Ensure:** Vacuum arm retraction command for iteration $i$
1: $t \leftarrow (i-1)/(n-1)$ *// Normalize iteration to [0,1]*
2: *// Incremental adjustments in vacuum arm base frame:*
3: $\Delta y \leftarrow 3.0 \cdot (1-\sqrt{t})$ mm *// Lateral: 3mm → 0mm*
4: $\Delta z \leftarrow 1.0 \cdot \sqrt{t}$ mm *// Vertical: 0mm → 1mm*
5: $\Delta\theta_x \leftarrow -5.0 \cdot (1-\sqrt{t})^\circ$ *// Tilt about base X-axis*
6: $\Delta z_{tool} \leftarrow -0.25 \cdot (1-\sqrt{t})$ mm *// Along tool Z-axis*
7: **return** Updated vacuum pose with cumulative adjustments

States 13-15 execute the shutdown sequence by retracting the electrosurgical arm, releasing vacuum grasping, and returning the vacuum arm to home position (State 1, Figure 9.a), readying the system for subsequent procedures.

## VI. Robotic Platform

We implement the Multi-shot workflow on an autonomous camera-guided dual-arm robotic platform composed of commercial components, custom tools, and a dedicated control stack, as shown in Figure 1. This platform is the execution substrate for the perception and planning pipeline described above rather than the primary methodological contribution.

### A. Hardware Architecture

The hardware system consists of two arms: a camera and electrosurgical cutting arm, and a vacuum-based grasping arm, both mounted on a hydraulic table and supported by auxiliary equipment. Vacuum grasping provides safer tissue manipulation than forceps, reducing tissue damage [54].

For future clinical translation, we select either FDA-approved equipment, or products from industry-leading manufacturers with existing FDA approvals in surgical robotics, such as UR manipulators (Universal Robots, Odense, Denmark) used in TMS-Cobot (Axilum Robotics, Strasbourg, France). The vacuum pads (VacuForce, Indianapolis, IN) are also FDA-compliant and widely used in the food industry.

The camera and cutting arm comprises a 6-degree of freedom (DoF) UR10e manipulator (Universal Robots), an RGB-D camera (2+ M60, Zivid, Oslo, Norway), a custom $350\,\mathrm{mm}$ laparoscopic extension, and a standard $25\,\mathrm{mm}$ long, $1\,\mathrm{mm}$ diameter electrode needle (Bovie, Clearwater, FL).

The grasping arm consists of a 6-DoF UR5 manipulator (Universal Robots), an electric vacuum gripper (EPick, Robotiq, Levis, Canada), an inline moisture trap, a custom $250\,\mathrm{mm}$ laparoscopic extension, and a bellows vacuum pad with swappable sizes for different tumors.

The auxiliary setup includes an electrosurgical unit (ESU) (ASG-300ESU, DRE Veterinary, Louisville, KY) with grounding pad, a smoke evacuator (Smoke Shark, Bovie), and an air purifier (GC Multigas, IQAir, Goldach, Switzerland) for electrosurgical odor mitigation.

### B. Software Architecture

The control software uses Robot Operating System 2 (ROS 2) Humble distribution on Ubuntu 22.04 and open-source libraries, facilitating fast development and flexibility for customization to specific surgical requirements.

System integration begins with establishing a unified coordinate system. A checkerboard placed near the target sample defines a shared *world* coordinate frame. Hand-eye calibrations align the camera frame and manipulator base frames with this *world* frame, enabling coordinated multi-arm operation and accurate camera-guided control.

The system supports both task-space and joint-space control of the manipulators. The trajectory planner uses the Orocos Real-Time Toolkit (RTT) for real-time execution, the Orocos Kinematics and Dynamics Library (KDL) for kinematics, and the Reflexxes Motion Library (RML) for real-time interpolation and synchronization. In task-space control, Cartesian position and quaternion orientation are time-synchronized, with linear translation interpolation and Spherical Linear Interpolation (SLERP) for orientation. The resulting pose sequence is then converted to joint angles via inverse kinematics to move the manipulator along the desired path. In joint-space control, all joints interpolate linearly with synchronized motion. The control loop runs at $500\,\mathrm{Hz}$ for UR10e and $125\,\mathrm{Hz}$ for UR5. Collision detection is handled by the Coal library and performs pairwise mesh checks across all links of both arms, the RGB-D camera, laparoscopic extensions, electrosurgical needle, and vacuum pad. The vacuum gripper is controlled through ROS 2 service calls. As our ESU is not programmable, it is manually enabled during operation.

### C. Safety Architecture

Our system implements four layered safety interlocks. (1) Hardware interlocks: An assistant maintains direct physical control of the URCap activation on the teach pendant and of the electrosurgical tool power cable. Disabling the URCap rejects all external commands while preserving the arms' native gravity compensation, so the tools remain statically held; unplugging the electrosurgical tool power cable from the ESU immediately terminates energy delivery. (2) Real-time deployer halt: The human supervisor terminates the Orocos RTT deployer from the Ubuntu computer; the UR driver's real-time watchdog then applies a controlled arm stop within one control period ($0.002\,\mathrm{s}$ for UR10e and $0.008\,\mathrm{s}$ for UR5). This halt operates below the workflow planner and therefore does not require the FSM to reach a safe transition. (3) Controller-level speed envelopes: RML clamps commanded motion to phase-specific envelopes. In task-space, these are $2\,\mathrm{mm/s}$ and $5\,°/\mathrm{s}$ during electrosurgical cutting and in-tissue motion, $5\,\mathrm{mm/s}$ and $5\,°/\mathrm{s}$ during near-tissue non-contact motion, and $60\,\mathrm{mm/s}$ and $15\,°/\mathrm{s}$ in free space. In joint-space, these are $5\,°/\mathrm{s}$, $10\,°/\mathrm{s}$, and $15\,°/\mathrm{s}$ respectively across all joints, with the $15\,°/\mathrm{s}$ envelope used for wrist-up and wrist-down pose transitions. The $2\,\mathrm{mm/s}$ cutting speed yields approximately a $3\,\mathrm{s}$ human supervisor response window across a $6\,\mathrm{mm}$ margin. (4) Supervisor plan approval: Every CSI and PDD plan is displayed in RViz2 and requires explicit supervisor approval before the cutting arm moves; rejection returns the planner to the scene capture state. This is the principal interlock against unintended electrosurgical cutting.

## VII. Experiments

We conduct two sets of experiments to evaluate our system's performance in intraoperative guidance and the downstream robotic tumor resection in an open-field setup that mimics OPN access.. The first experiment set assesses the quality of occupancy-guided intraoperative inference by comparing our method's outputs against ground truth CT scans for four different patient kidney phantoms. The second experiment set evaluates the system's performance in autonomous tumor removal.

### A. Phantom Fabrication and Setup

In surgical oncology research, patient-derived 1:1 scale phantoms are clinically appropriate for surgical planning and rehearsal [55]. While silicone phantoms have been used [56], hydrogel phantoms offer electrical conductivity enabling electrosurgical cutting [41], [55]. Our patient CT-derived hydrogel phantom follows the design from Melnyk et al. [41], with added CT contrast agents for preoperative and postoperative imaging. *Ex vivo* kidneys present significant challenges: once removed from the body, kidneys lose perfusion pressure and collapse, and existing tumor mimic methods produce volumes too small for realistic resection [39], [40], precluding incorporation of patient-specific geometry from clinical CT scans.

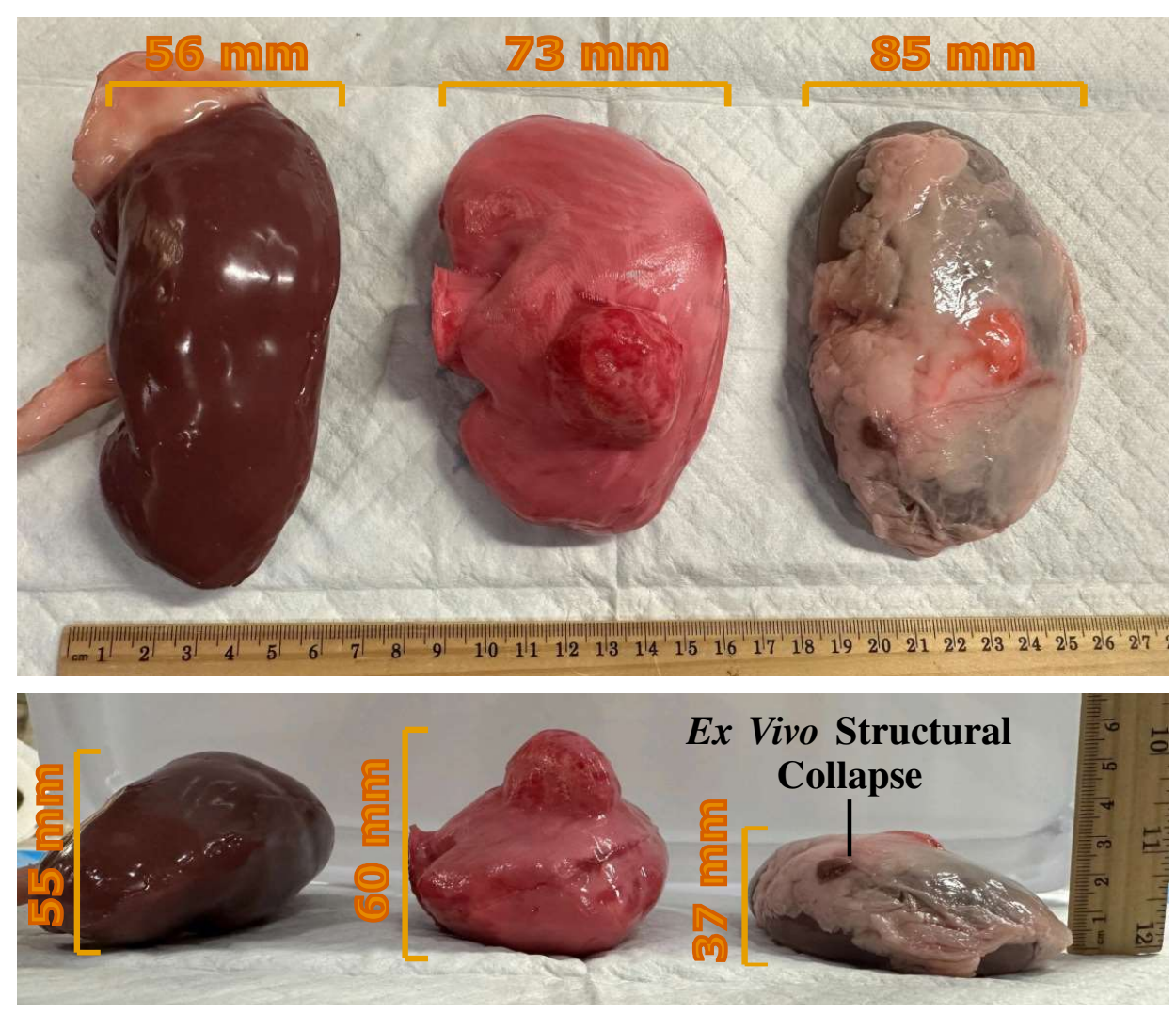

Fig. 10. Comparison of kidney models. Top: top view; Bottom: front view. Left: commercial silicone phantom with adrenal gland but no tumor (SynDaver); silicone lacks electrical conductivity required for electrosurgery. Middle: patient-derived hydrogel phantom with tumor [41]; hydrogel is electrically conductive and maintains patient-specific geometry. Right: *ex vivo* porcine kidney with tumor mimic [40]; collapsed renal cavity prevents realistic geometry, and parenchyma tearing limits tumor mimic to sub-clinical sizes.

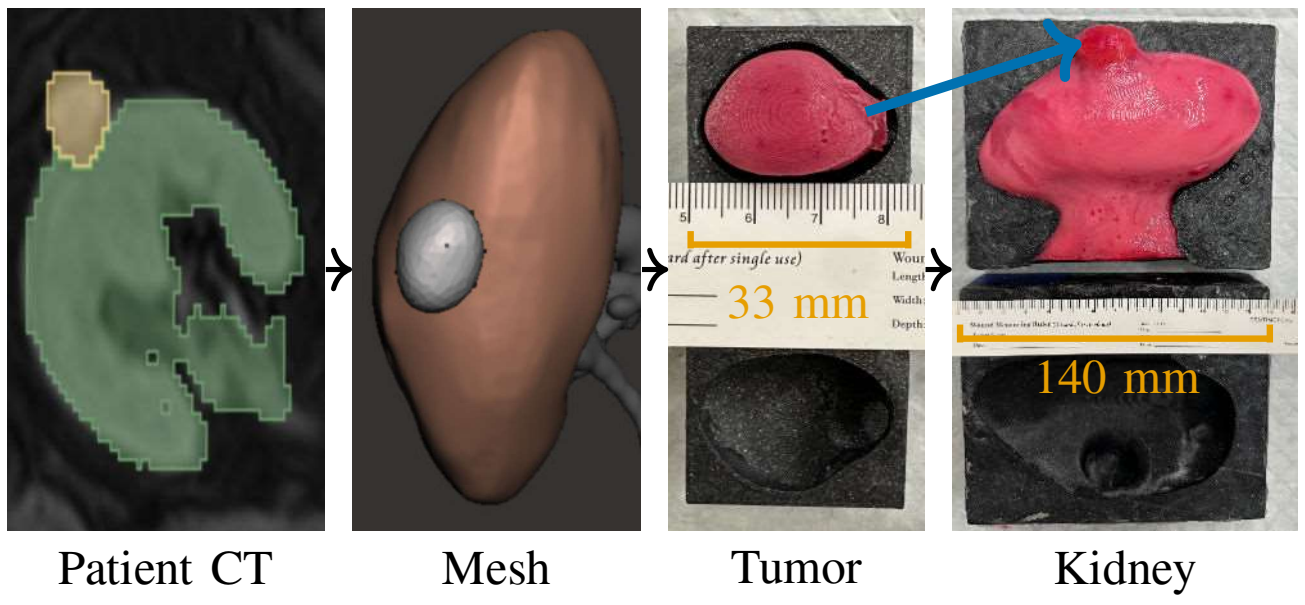

Fig. 11. Patient CT scans are used to generate volumetric meshes. The negative of each mesh is 3D-printed to create molds, which are then used to cast hydrogel kidney phantoms. A tumor model, with enhanced CT contrast for both preoperative and postoperative imaging, is prepared separately and inserted into the mold before fabricating the kidney phantom.

Figure 10 compares a commercial silicone kidney phantom with adrenal gland but no tumor (SynDaver, Tampa, FL), a patient-derived hydrogel phantom with tumor [41], and an *ex vivo* porcine kidney with tumor mimic [40]. Top and front views visualize the geometrical limitations of *ex vivo* models: collapsed renal cavity prevents realistic geometry, and limited tumor mimic volume before parenchymal tearing.

Following Melnyk et al. [41], our phantoms are made of hydrogel mixtures of water and polyvinyl alcohol (PVA) that stiffen into an electrically conductive material through freeze-thaw cycling, with mechanical properties controlled through cycle count and PVA concentration, validated to mimic porcine kidney tissue. Figure 11 shows the phantom creation process.

We obtain anonymized patient CT data of PN cases featuring kidneys with partially exophytic tumors. The data is segmented into tumor and kidney and converted to surface meshes. Negative mold meshes, produced using boolean dif-ference operations, are 3-D printed in polylactic acid (PLA) on an Ender-3 Max Neo (Creality, Shenzhen, China). Kidney molds and matching tumor molds are printed for each patient. Two separate batches of 10% w/v PVA are prepared, one for the kidney and one for the tumor. The tumor batch is mixed with an additional 2% w/v barium sulfate (Sigma-Aldrich, St. Louis, MO) to improve CT contrast and enable preoperative and postoperative segmentation. Food coloring is also added to both batches for a red or yellow appearance. The tumor PVA mixture is injected into the tumor mold and undergoes one freeze-thaw cycle. After thawing, the tumor is positioned into the correct placement within the kidney mold before filling with the kidney PVA mixture. The combined tumor-containing kidney undergoes one final freeze-thaw cycle, completing the phantom. In practice, we use four patient kidney CT scans and refer to each distinct kidney as a *patient model*, visualized in Figure 12. For the experiments, we also prepare thick hydrogel pads for the kidney phantoms as sample holders.

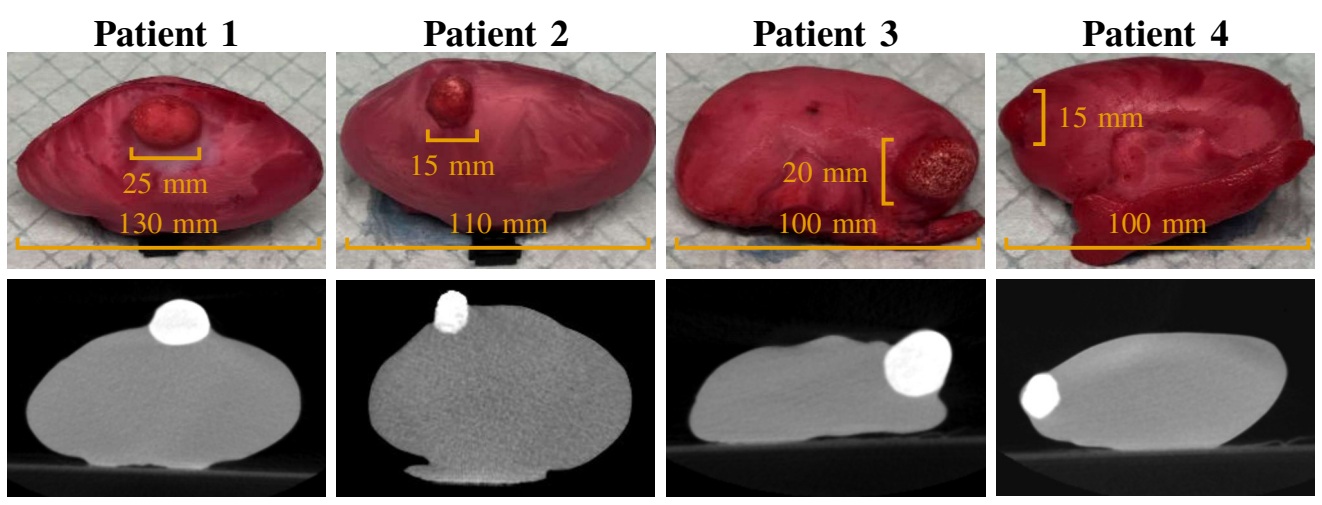

Fig. 12. Four distinct kidney and tumor models are obtained from clinical partial nephrectomy cases and referred to as *patient models*. Preoperative CT scans are taken of all patients and used for postoperative analysis.

### B. Experimental Design: 3-D Anatomy Inference

We manufacture one hydrogel phantom of each patient model (4 in total) according to Section VII-A. Additionally, we train occupancy networks for each patient model according to Section IV-A. We glue the hydrogel phantom into a sample holder, which is placed on an electrosurgical grounding pad. Note, the sample holder is not fixated and is free to move on the table as forces are applied. We then cover the surgical site using drapes, mimicking an open surgical setting [25], [26]. The fixed and draped phantom model is then placed within a Loop-X (Brainlab, Munich, Germany) for acquiring CT scans of the kidney phantom across resection states. Manual electrosurgical cutting and vacuum gripper manipulation are used to perform the tumor resection. The laparoscopic extension to the vacuum gripper is swapped for a polycarbonate material to avoid artifacts caused by inserting metal into the CT scanner. The Zivid camera is mounted to a fixed observation of the scene, as shown in Figure 13. Paired CT and depth point cloud data is collected across five approximate resection states: pre-CSI, 0% PDD, 25% PDD, 50% PDD, and 75% PDD (example scenes shown in Figure 15), for a total of 20 data pairs. CT scans are segmented into kidney and tumor, and registered to the Zivid coordinate frame using manual landmark-based registration, followed by ICP for fine adjustment. The resulting dataset enables comparison between ground truth CT volumes and occupancy network predictions across resection states.

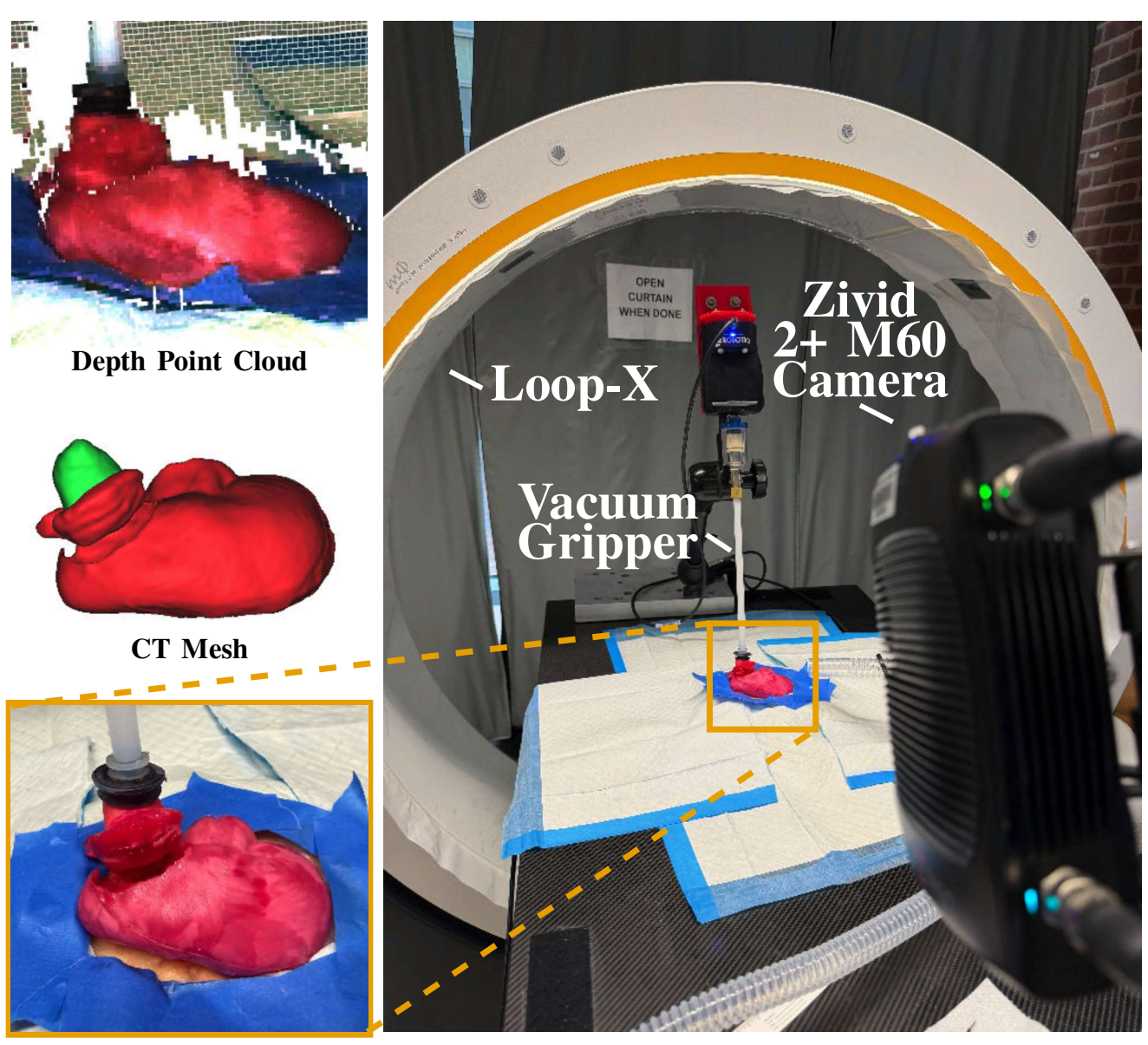

Fig. 13. Experimental setup for paired CT and depth point cloud data collection. A hydrogel kidney phantom is placed within a Loop-X CT scanner while a Zivid 2+ M60 depth camera observes the scene from a fixed position. At each of five resection states (CSI, 0% PDD, 25% PDD, 50% PDD, and 75% PDD), paired CT and depth point cloud data are recorded, yielding 20 data pairs across four patient models. The CT volume is registered to the depth point cloud coordinate frame and serves as a reference for evaluating occupancy network predictions.

**CSI 3-D Anatomy Inference:** We compare the tumor localization accuracy of the CSI-phase occupancy network to ICP, CPD, and the deformable registration approach Volume2SurfaceCNN (V2S) [32]. While a comparison of the circumferential cut would be more suitable, there is no ground-truth label provided by the CT. For ICP, CPD, and V2S we register the preoperative 3-D mesh (see Figure 5) to the Zivid point cloud. The aligned model is then compared to the CT ground truth to assess tumor localization accuracy. We provide both ICP, CPD and V2S with an approximate manual rigid alignment, as we found both not to converge to reasonable solutions without.

**PDD 3-D Anatomy Inference:** We compare the accuracy of the PDD-phase occupancy network to a rigid-assumption baseline. The rigid approach begins with the CSI occupancy network predictions for the kidney, tumor, and margin. The rigid assumption holds that the shapes and sizes of the tumor and margin tissue remain unchanged throughout resection, and that both move rigidly with the vacuum gripper. The motion of the vacuum gripper between resection states is recovered by aligning the vacuum cup model from the prior resection state CT to its position in the current CT volume using ICP. The resulting transform is then applied to the rigid tumor and margin tissue to estimate their pose at each subsequent resection state.

We also evaluate the effectiveness of the cut surface randomization introduced in Section IV-C. For this, we train a PDD-phase model for each patient without surface randomization.

### C. Evaluation Metrics: 3-D Anatomy Inference

**Optimal Margin Identification:** A successful occupancy network prediction identifies the appropriate target margin to resect, defined as $6\,\mathrm{mm}$ from the ground truth CT tumor. In the occupancy network prediction, the target is the intersecting surface of the inferred margin and kidney, shown as the optimal margin in Figure 6. To evaluate accuracy, points are sampled uniformly on the occupancy-predicted optimal margin and, for each sampled point, the distance to the closest point on the CT tumor is computed. We choose to compute this metric relative to the registered CT segmentation of the tumor because our phantom model does not have any distinguishable contrast for the desired resection margin. We report the absolute and signed error relative to the $6\,\mathrm{mm}$ desired margin distance to the tumor.

**Tumor Localization:** To evaluate the tumor localization accuracy, we report the centroid error, 95% Hausdorff distance, and intersection over union (IoU) between the occupancy-predicted tumor and the CT tumor. All metrics are reported per patient model, and as averages across all.

Statistical significance of PDD performance differences between the occupancy network and the rigid-assumption baseline is assessed using a paired t-test across the $N = 16$ evaluation scenes. Optimal margin identification comparisons of absolute and signed error are performed on per scene average values. $P < 0.05$ is considered significant. The CSI phase is not statistically analyzed due to the low sample size.

### D. Experimental Design: Robotic Resection

We manufacture four hydrogel phantoms of each patient model (16 in total) according to Section VII-A. For each hydrogel phantom, we acquire a CT image prior to the experiments. The trained occupancy networks and phantom setup are used from the experiments described in Section VII-B. The power of the electrosurgical instrument is set to $30\,\mathrm{W}$ for all experiments. Two bellows vacuum pads with different inner diameter (ID) and outer diameter (OD) are used to adapt to varying tumor sizes. For patient 1 and 3, a pad with an ID of $11\,\mathrm{mm}$ and an OD of $24\,\mathrm{mm}$ is used. For patient 2 and 4, a pad with an ID of $10\,\mathrm{mm}$ and an OD of $21\,\mathrm{mm}$ is used.

**Multi-shot Occupancy-Guided Workflow (Main Method)**: The robotic platform executes the proposed Multi-shot occupancy-guided workflow as outlined in Section V. The target being a $6\,\mathrm{mm}$ tissue margin. We perform two human supervised PN for each patient model.

**Single-shot Rigid-Assumption Baseline**: To evaluate the advantage of iterative Multi-shot guidance, we compare it to a Single-shot baseline. This approach begins with a single-time point cloud capture and full 3-D kidney anatomy inference, and followed by CSI planning, execution, and vacuum-based tumor grasping, consistent with States 1-10 in Figure 9 for the Multi-shot method. Afterward, the Single-shot method assumes that the shapes and sizes of the tumor and margin tissue remain unchanged and that both move rigidly with the vacuum pad during resection. Consequently, the location of the tumor and margin tissue can be calculated using forward kinematics based on the pose of the vacuum gripper frame, as shown in Figure 9.a. Unlike the Multi-shot method, which requires iterative point cloud captures and occupancy inferences during PDD planning, the Single-shot method operates solely within States 11-12, bypassing States 2-4 needed by the Multi-shot method. Therefore, this baseline only uses the preoperative CT and a single initial point cloud observation for guidance during cutting (no perception updates). All remaining procedures are identical to those described in Section V. The target is a $6\,\mathrm{mm}$ tissue margin. We perform one human supervised PN for each patient model.

**Surgeon**: We invite a board-certified PN (urologic) surgeon to perform resections as a human surgeon comparison. The surgeon reviews the preoperative CT scan of each patient prior to each procedure. The surgeon is instructed to resect with the same workflow and margin sizes that they would use for a real patient. Following their preferences, the surgeon chooses to make surface marks using a handheld electrosurgical pen with an electrode spatula (Bovie) and deep dissection cuts with a surgical scissors. Any necessary grasping was performed with a delicate handheld grasp by surgical gauze. The surgeon performs one PN for each patient model.

### E. Evaluation Metrics: Robotic Resection

**Tumor Removal Success:** A tumor removal is considered successful, in our evaluation, if no contrast material is found on the margin outer surface of the removed mass in the postoperative CT. This is analogous to the clinical definition of complete resection with an NSM, where no tumor cells are found at the surgical margin boundary. The success rate is reported for each study group, an example of a failed case is shown in Figure 16.D, indicating two PSM sites.

**Margin Accuracy and Consistency:** The margin size is defined as the distance between the tumor surface and the margin outer surface, as shown in Figure 2. The resection surface is extracted as shown in Figure 14 and we sample $100$ points per $\mathrm{mm}^2$ uniformly on its surface. This sampling density was chosen to ensure stable metrics calculation over different point sample initiations. For each sampled point, the closest point on the tumor surface is determined, providing a per-point margin distance. Using these distances, we compute the mean margin size, its standard deviation, and the minimum margin size, which corresponds to the closest cut to the tumor. From these, we derive the coefficient of variation (COV), defined as the ratio between standard deviation and mean margin (expressed as a percentage where lower values are better), to quantify resection consistency. For the Multi-shot and Single-shot groups, absolute and signed margin errors are also reported relative to the $6\,\mathrm{mm}$ margin target. Since the Surgeon group did not aim for a specific margin size, error metrics are not reported.

**Resection Volume and Duration:** The volume of removed kidney tissue is estimated by subtracting the postoperative kidney volume from the preoperative volume, accounting for tissue that was vaporized or otherwise not captured in the excised specimen. Additionally, the average procedure duration is reported for all study groups, and the mean number of supervised re-plans is provided for the Multi-shot group.

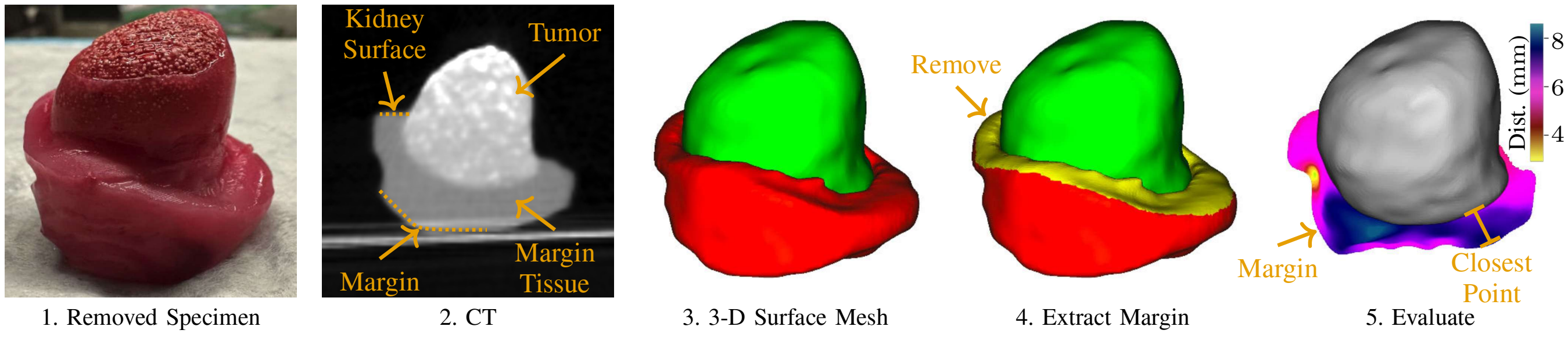

Fig. 14. Process to evaluate the robotic resection. After the surgical removal, the specimen is imaged in a CT and segmented. The segmented CT is converted to 3-D surface meshes. The margin is obtained from the margin tissue by removing the surfaces that were the kidney surface (surfaces perpendicular to nearby tumor surface) and in contact with the tumor. For each point on the margin, the distance to the tumor is computed (closest point on the tumor).

## VIII. Results

### A. 3-D Anatomy Inference

TABLE II
TUMOR LOCALIZATION ERROR DURING THE CSI PHASE. $N = 4$ SAMPLES (ONE FOR EACH PATIENT). CENTROID ERROR, HAUSDORFF 95% DISTANCE, AND IOU FOR ICP, CPD, V2S, AND CSI-PHASE OCCUPANCY NETWORK.

| Patient | Centroid Error (mm) | | | | Hausdorff 95% (mm) | | | | IoU | | | |
|---|---|---|---|---|---|---|---|---|---|---|---|---|
| | ICP | CPD | V2S | CSI-Occ. | ICP | CPD | V2S | CSI-Occ. | ICP | CPD | V2S | CSI-Occ. |
| 1 | 3.56 | 5.73 | 6.36 | **2.42** | 4.59 | 7.93 | 8.02 | **4.37** | 0.54 | 0.42 | 0.38 | **0.67** |
| 2 | 7.25 | **2.40** | 5.97 | 3.55 | 7.04 | **2.87** | 6.87 | 3.23 | 0.27 | **0.62** | 0.33 | 0.50 |
| 3 | 8.79 | 3.16 | 2.09 | **1.98** | 9.17 | **3.85** | 6.03 | 4.04 | 0.28 | **0.68** | 0.67 | 0.56 |
| 4 | 7.21 | 7.16 | 7.86 | **4.37** | 7.44 | 7.38 | 7.75 | **2.77** | 0.26 | 0.27 | 0.22 | **0.52** |
| Average | 6.70 $\pm 1.92$ | 4.61 $\pm 1.92$ | 5.57 $\pm 2.13$ | **3.08** $\pm \mathbf{0.94}$ | 7.06 $\pm 1.64$ | 5.51 $\pm 2.18$ | 7.17 $\pm 0.78$ | **3.60** $\pm \mathbf{0.63}$ | 0.34 $\pm 0.12$ | 0.50 $\pm 0.16$ | 0.40 $\pm 0.17$ | **0.56** $\pm \mathbf{0.07}$ |

Representative results of the 3-D anatomy inference experiments are visualized in Figure 15.

**CSI 3-D Anatomy Inference:** Table II summarizes the results of the 3-D anatomy inference experiments on the CSI phase. For centroid localization, the CSI occupancy network achieved the lowest average error with $3.08 \pm 0.94$ mm, compared with $6.70 \pm 1.92$ mm for ICP, $4.61 \pm 1.92$ mm for CPD, and $5.57 \pm 2.13$ mm for V2S. The CSI occupancy network also performed best in terms of Hausdorff 95% distance, with $3.60 \pm 0.63$ mm compared with $7.06 \pm 1.64$ mm for ICP, $5.51 \pm 2.18$ mm for CPD, and $7.17 \pm 0.78$ mm for V2S. Similarly, it achieved the highest average IoU with $0.56 \pm 0.07$, compared with $0.34 \pm 0.12$ for ICP, $0.50 \pm 0.16$ for CPD, and $0.40 \pm 0.17$ for V2S.

**PDD 3-D Anatomy Inference:** Table III summarizes the results of the inference quality experiments on the PDD phase. Across four phantoms and 16 total PDD resection states, the average absolute margin identification error for the PDD occupancy networks was $1.97 \pm 1.55$ mm, and the average signed margin identification error was $0.83 \pm 2.36$ mm. The average centroid error was $2.13 \pm 1.04$ mm, the average Hausdorff 95% distance was $3.30 \pm 1.25$ mm, and the average IoU was $0.66 \pm 0.08$.

For the rigid comparison, the average absolute margin identification error was $2.13 \pm 1.62$ mm, and the average signed margin identification error was $0.34 \pm 2.66$ mm. The average centroid error was $6.29 \pm 2.28$ mm, average Hausdorff 95% distance was $6.70 \pm 2.35$ mm, and average IoU was $0.35 \pm 0.11$.

Differences between the PDD occupancy network and rigid-assumption were significant for absolute margin identification error ($P < 0.05$), centroid error ($P < 0.001$), Hausdorff 95% ($P < 0.001$), and IoU ($P < 0.001$). The differences between the two methods for signed margin error was not significant ($P = 0.796$).

Not randomizing the cut surface during training results in an average tumor localization error of $2.35 \pm 0.85$ mm, Hausdorff 95% of $3.39 \pm 1.21$ mm, and IoU of $0.65 \pm 0.06$. More detailed results are presented in Table VIII (see the Appendix).

### B. Robotic Resection

Results of all 16 resection experiments are visualized in Figure 16. Table IV presents resection accuracy using all sampled margin points from the CT-derived post-resection margin outer surfaces, Table V reports metrics averaged per resected kidney tumor, and Table VI details per-procedure metrics for Multi-shot experiments including cuts, re-plans, and duration.

**Multi-shot (Main Method)**: In eight experiments, all tumor resections resulted in NSMs, and average margin size was $6.63 \pm 1.88$ mm. Consistency measured by the COV was 28.31% and the average minimum margin size was $2.17 \pm 0.92$ mm. Compared to the 6 mm target, the average absolute margin error was $1.63 \pm 1.13$ mm, and the average signed error was $+0.63 \pm 1.88$ mm (positive indicates larger-than-planned margin size). The average volume of kidney tissue removed was $11631.54\,\text{mm}^3$, the average procedure duration was 12 min and 2.08 s.

**Single-shot**: In four experiments, the tumor resection success rate was 0% (three because of tumor puncture and one aborted experiment due to unacceptable tool proximity). The average margin size was $3.80 \pm 1.78$ mm. Consistency measured by the COV was 46.93%. Since every resection failed, and the penetration distance is irretrievable with the postoperative CT, the minimum margin size was unattainable. Compared to the 6 mm target, the average absolute margin error was $2.38 \pm 1.53$ mm, and the average signed error was $-2.20 \pm 1.78$ mm. The average volume of kidney tissue removed was $7428.62\,\text{mm}^3$, and the average procedure duration was 6 min and 30.74 s.

**Supervisor Monitoring Log**: Tool-to-tool or tool-to-robot collisions: 0 for both Multi-shot and Single-shot. Unintended ESU activations or emergency disconnections: 0 for both. Emergency halts: 0 and 1, respectively (the Single-shot halt

TABLE III

OPTIMAL MARGIN IDENTIFICATION AND TUMOR LOCALIZATION ERROR DURING THE PDD PHASE. MARGIN IDENTIFICATION EVALUATES THE METHOD'S ABILITY TO IDENTIFY THE APPROPRIATE MARGIN FOR RESECTION AS ABSOLUTE AND SIGNED ERROR FROM DESIRED 6 mm. CENTROID ERROR, HAUSDORFF 95%, AND IOU FOR TUMOR LOCALIZATION. N=16 SAMPLES (FOUR FOR EACH OF THE FOUR PATIENTS). RIGID REFERS TO THE RIGID-ASSUMPTION BASELINE (SEE SECTION VII-B).

| Patient | Optimal Margin Identification | | | | Tumor Localization | | | | | |
|---|---|---|---|---|---|---|---|---|---|---|
| | Abs. Error (mm) | | Sig. Error (mm) | | Centroid Error (mm) | | Hausdorff 95% (mm) | | IoU | |
| | PDD-Occ. | Rigid | PDD-Occ. | Rigid | PDD-Occ. | Rigid | PDD-Occ | Rigid | PDD-Occ | Rigid |
| 1 | $2.70 \pm 2.02$ | $\mathbf{2.28 \pm 1.96}$ | $+2.28 \pm 2.48$ | $\mathbf{+0.83 \pm 3.24}$ | $\mathbf{2.53 \pm 1.31}$ | $9.48 \pm 0.51$ | $\mathbf{3.52 \pm 1.30}$ | $10.20 \pm 0.71$ | $\mathbf{0.62 \pm 0.10}$ | $0.37 \pm 0.02$ |
| 2 | $\mathbf{1.14 \pm 0.71}$ | $2.85 \pm 1.82$ | $\mathbf{+0.73 \pm 1.13}$ | $+2.69 \pm 2.04$ | $\mathbf{2.14 \pm 0.36}$ | $4.45 \pm 1.05$ | $\mathbf{2.47 \pm 0.29}$ | $4.58 \pm 1.04$ | $\mathbf{0.68 \pm 0.03}$ | $0.35 \pm 0.08$ |
| 3 | $\mathbf{2.34 \pm 1.54}$ | $3.15 \pm 1.92$ | $\mathbf{+1.89 \pm 2.07}$ | $+2.10 \pm 3.03$ | $\mathbf{2.54 \pm 1.37}$ | $6.83 \pm 1.10$ | $\mathbf{4.70 \pm 1.24}$ | $6.83 \pm 1.10$ | $\mathbf{0.67 \pm 0.10}$ | $0.42 \pm 0.07$ |
| 4 | $\mathbf{1.72 \pm 1.08}$ | $2.37 \pm 1.65$ | $\mathbf{-1.57 \pm 1.28}$ | $-1.67 \pm 2.35$ | $\mathbf{1.32 \pm 0.57}$ | $4.39 \pm 0.42$ | $\mathbf{2.52 \pm 0.39}$ | $5.25 \pm 0.47$ | $\mathbf{0.67 \pm 0.08}$ | $0.19 \pm 0.04$ |
| Average | $\mathbf{1.97 \pm 1.55}$ | $2.13 \pm 1.62$ | $+0.83 \pm 2.36$ | $\mathbf{+0.34 \pm 2.66}$ | $\mathbf{2.13 \pm 1.04}$ | $6.29 \pm 2.28$ | $\mathbf{3.30 \pm 1.25}$ | $6.70 \pm 2.35$ | $\mathbf{0.66 \pm 0.08}$ | $0.35 \pm 0.11$ |

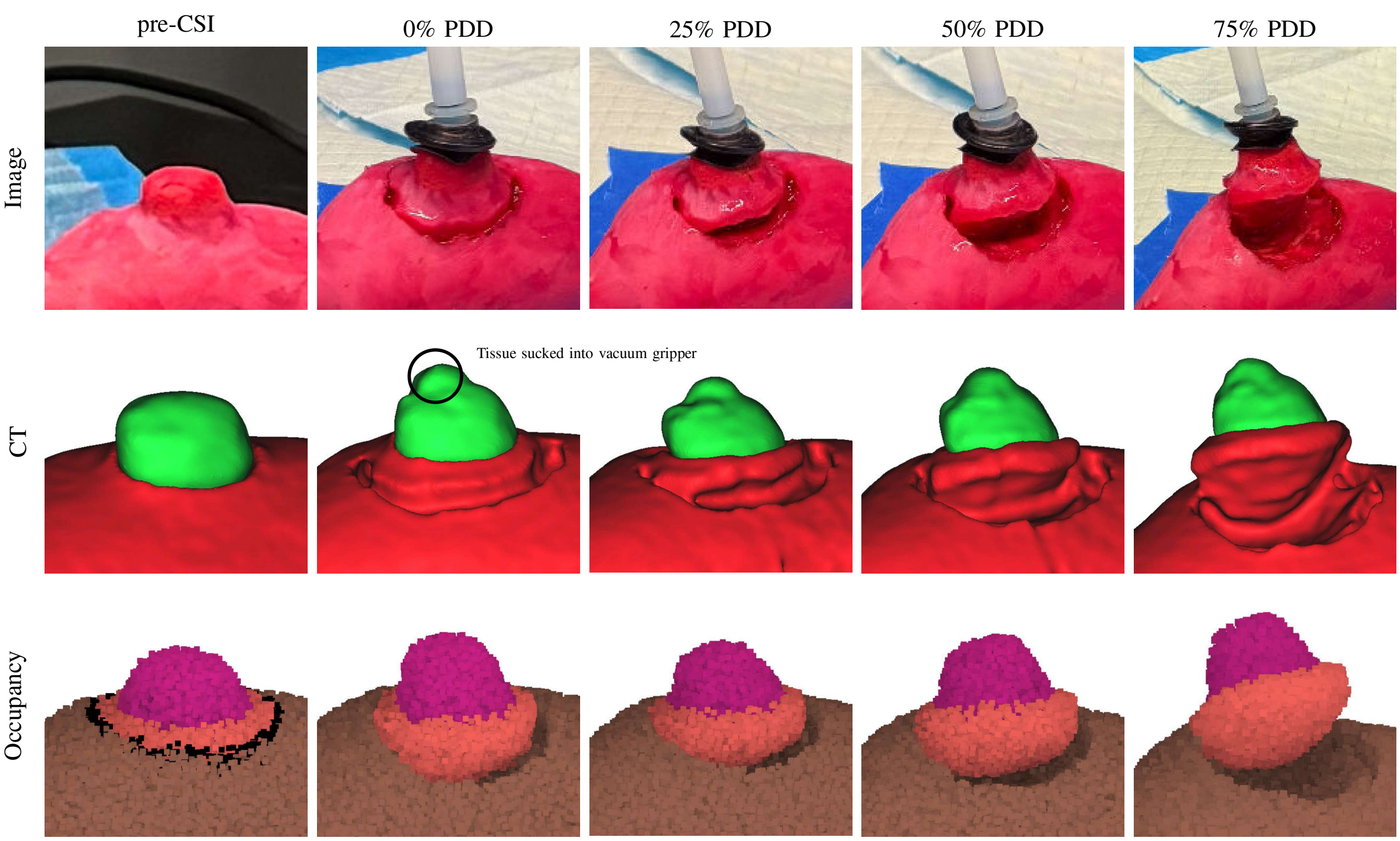

Fig. 15. Representative real-world reference CT data and inferences using the CSI and PDD occupancy networks during various stages of resection. A noticeable difference between the occupancy and the CT data is the small area sucked up into the vacuum gripper. The training data of the occupancy network did not include this behavior.

TABLE IV

RESECTION ACCURACY RESULTS FOR EACH METHOD, CALCULATED FROM ALL SAMPLED MARGIN POINTS ON CT-DERIVED POST-RESECTION MARGIN OUTER SURFACES. SAMPLE SIZE (N) FOR EACH METHOD REFLECTS THE NUMBER OF TUMORS RESECTED AND THE SURFACE AREA OF THE CORRESPONDING MARGINS. MARGINS AND ERRORS ARE REPORTED AS MEAN ± STANDARD DEVIATION. ERROR VALUES ARE OMITTED FOR THE SURGEON GROUP, WHICH DID NOT TARGET A SPECIFIC MARGIN SIZE. BEST RESULTS SHOWN IN BOLD.

| Metric<br>Sampled Margin Points | **Multi-shot**<br>N=970823 | **Single-shot**<br>N=274387 | **Surgeon**<br>N=387079 |
|---|---|---|---|
| Margin Size (mm) | 6.63±1.88 | 3.80±1.78 | 4.47±2.63 |
| Abs. Error (mm) | **1.63±1.13** | 2.38±1.53 | - |
| Sig. Error (mm) | **+0.63±1.88** | -2.20±1.78 | - |
| COV (std/mean%) | **28.31** | 46.93 | 58.82 |

TABLE V

RESECTION ACCURACY RESULTS FOR EACH METHOD, CALCULATED BY AVERAGING OVER INDIVIDUAL RESECTED TUMORS. SAMPLE SIZE (N) FOR EACH METHOD REFLECTS THE NUMBER OF TUMORS RESECTED. MARGINS, ERRORS, AND COV ARE REPORTED AS MEAN ± STANDARD DEVIATION. ERROR VALUES ARE OMITTED FOR THE SURGEON GROUP, WHICH DID NOT TARGET A SPECIFIC MARGIN SIZE. BEST RESULTS SHOWN IN BOLD.

| Metric<br>Resected Tumors | **Multi-shot**<br>N=8 | **Single-shot**<br>N=3 | **Surgeon**<br>N=4 |
|---|---|---|---|
| Success (%) | **100** | 0 | **100** |
| Margin Size (mm) | 6.41±1.04 | 3.78±0.68 | 4.25±1.49 |
| Margin Min (mm) | 2.17±0.92 | - | 0.57±0.16 |
| Abs. Error (mm) | **1.61±0.48** | 2.42±0.56 | - |
| Sig. Error (mm) | **+0.41±1.04** | -2.22±0.68 | - |
| COV (std/mean%) | **25.83 ± 10.52** | 44.64 ± 4.82 | 42.06 ± 7.55 |

TABLE VI

MULTI-SHOT (MS) PROCEDURAL METRICS FOR N=8 EXPERIMENTS ACROSS 4 PATIENT MODELS (PAT.). EACH PROCEDURE (PROC.) CONSISTS OF ONE CSI CUT FOLLOWED BY MULTIPLE PDD CUTS. A PROCEDURE ACHIEVES AN NSM ONLY IF ALL CUTS SUCCEED; A SINGLE FAILED CUT RESULTS IN A PSM AND PROCEDURE FAILURE. ALL 77 CUTS WITH 20 RE-PLANS ACROSS 8 PROCEDURES ACHIEVED NSMS.

| Proc. | Pat. | CSI+PDD Re-plans | CSI+PDD Cuts | Cut Succ. | Dur. (s) | NSM |
|---|---|---|---|---|---|---|
| Ms 1 | 1 | 0+1 | 1+10 | 11/11 | 715.21 | ✓ |
| Ms 2 | 1 | 0+0 | 1+10 | 11/11 | 595.61 | ✓ |
| Ms 3 | 2 | 0+1 | 1+8 | 9/9 | 592.91 | ✓ |
| Ms 4 | 2 | 1+7 | 1+8 | 9/9 | 1057.21 | ✓ |
| Ms 5 | 3 | 0+3 | 1+8 | 9/9 | 718.51 | ✓ |
| Ms 6 | 3 | 0+4 | 1+9 | 10/10 | 916.81 | ✓ |
| Ms 7 | 4 | 0+0 | 1+9 | 10/10 | 616.91 | ✓ |
| Ms 8 | 4 | 0+3 | 1+7 | 8/8 | 563.50 | ✓ |
| Total | 4 | 20 | 77 | 77/77 | 5776.67 | 8/8 |
| Avg. | – | 2.50 | 9.63 | – | 722.08 | 100% |

caused the aborted Patient 2 trial in Figure 16). Vacuum-grasp failures: 3 and 0, respectively. Supervisor re-plan requests: 20 total (average 2.50 per procedure) and 0, respectively. Note, Single-shot does not re-capture data during the procedure; therefore, as long as the initial capture is adequate, no supervisor re-plans can be performed.

**Surgeon**: In four experiments, all tumor resections resulted in NSMs, and the average margin size was $4.47 \pm 2.63\,\mathrm{mm}$. Consistency measured by the COV was $58.82\%$ and the average minimum margin size was $0.57 \pm 0.16\,\mathrm{mm}$. The average volume of healthy kidney removed was $5245.75\,\mathrm{mm}^3$, and the average procedure duration was $3\,\mathrm{min}$ and $46\,\mathrm{s}$.

## IX. DISCUSSION

### A. 3-D Anatomy Inference

**CSI 3-D Anatomy Inference:** The CSI-phase occupancy network consistently outperforms, ICP, CPD, and V2s (see Table II) with mean centroid errors of $3.08 \pm 0.94\,\mathrm{mm}$, $6.70 \pm 1.92\,\mathrm{mm}$, $4.61 \pm 1.92\,\mathrm{mm}$, and $5.57 \pm 2.13\,\mathrm{mm}$ respectively. Still, the error of the CSI-phase occupancy network is larger than expected. The inaccurate estimation of the tumor depth is a contributor to this error. This inaccurate estimate however, does not affect the CSI-phase in a meaningful way, as the circumferential cut is performed $10\,\mathrm{mm}$ deep and small deviations are unlikely to cause resection issues. Still, positional errors will propagate to the PDD-phase.

We observed that ICP and CPD share a common failure mode in our setting. The kidney surface is largely smooth and ellipsoidal, while the tumor often manifests as a small, localized protrusion represented by only a limited number of points in the observed point cloud. Consequently, the registration objective becomes highly underconstrained: large portions of the observation can be "slid" along the organ surface with minimal change in the optimization cost. This is further exacerbated by the need for distance thresholds during optimization for partial-observation registration. Without distance thresholding, the found solution would always place the to-be-registered 3-D object centered on the observation (geometrically the closest solution). However, this distance thresholding results in the tumor points (the part of the observation showing the tumor) being easily ignored. Further, we observed that the exophytic tumors are not well-represented in the sparse voxel ($64^3$) representation used by V2S. We found that, ICP, CPD, and V2S require an initial alignment to find a reasonable solution. Note, unlike for the CT to Zivid point cloud registration, no manual landmark-based registration was performed. While weighting methods could potentially improve ICP and CPD, automatically identifying such features in sparse observations is a challenge beyond the scope of this work. Alternatively, using a feature-based or globally optimal registration approach such as Go-ICP [57] could provide improvement at the cost of computation time.

In contrast, the occupancy network learns to focus on semantically important structures such as the tumor [36], enabling reliable localization without explicit correspondence matching or pre-alignment.

**PDD 3-D Anatomy Inference:** The PDD-phase occupancy network outperforms the rigid-assumption comparison method and achieves statistical significance on absolute margin identification error, centroid error, and Hausdorff 95% distance. While statistical significance was not achieved for signed margin identification error, we consider this metric primarily as an indicator of over- or under-estimating the margin for resection, and less useful in overall accuracy analysis.

A notable observation from the captured CT data reveals that the tumors are drawn into the vacuum cup during retraction. This real world deformation was not present in the simulation training data and likely contributed to variations in both centroid error and Hausdorff distance measurements. Patient 4 shows this effect most clearly, where the tumor's narrow morphology allowed substantial deformation within the cup, resulting in higher-than-average Hausdorff distances.

In our evaluation, we sought to emphasize our system's ability to localize the tumor for resection. Centroid error and Hausdorff distance both provide worst-case analyses of tumor localization, and IoU provides an analysis of accurate volume placement, all clinically relevant for avoiding PSM and resection failure. The absolute and signed margin identification errors offer meaningful insights for the downstream robotic resection. These metrics report the distance between the occupancy predicted margin and tumor as error from the $6\,\mathrm{mm}$ margin target. While reporting the error relative to the ground truth would have been more suitable, this margin is not observable in the CT scans. However, our occupancy network tracks the deforming margin from the pre-resection state, revealing a limitation that this metric does not account for the margin deformation.

The intuition behind providing the rigid-assumption baseline for localizing the tumor was that knowing the rigid motion of the vacuum gripper is enough for accurate localization. In practice, we observed substantial tumor lifting and deformation immediately upon vacuum activation. Since deformation occurs before the vacuum gripper moves, the rigid baseline is unable to account for this, and localization errors propagate throughout the resection. In contrast, the occupancy network maintains accurate tumor localization despite ongoing defor-mation and tissue removal. Standard deformable registration assumes fixed topology and is therefore inappropriate here, because the tumor–margin region is directly affected by cutting. An online FEM baseline would likewise need real-time topological updates and sim-to-real matching, which is beyond a baseline comparison for this study.

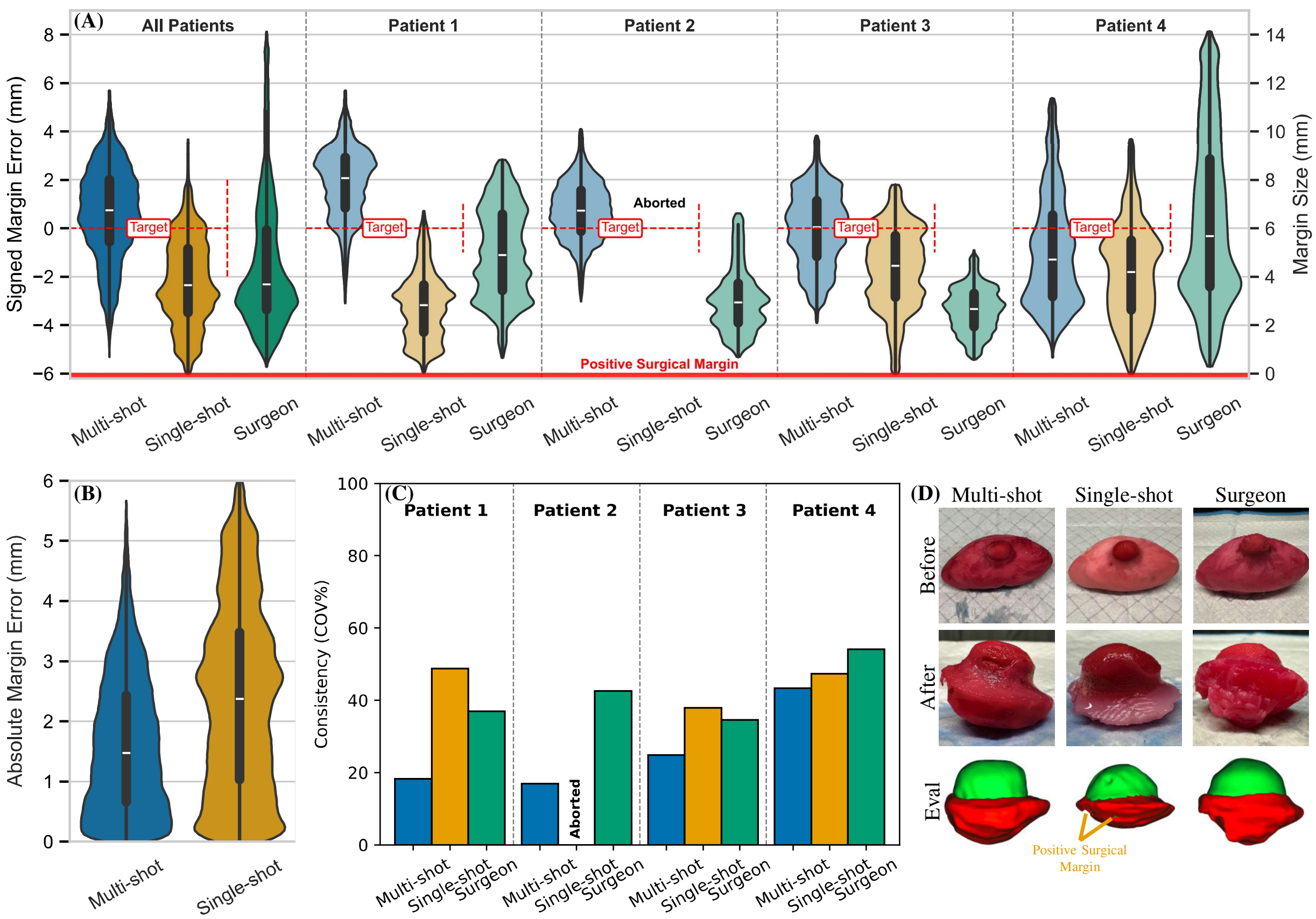

Fig. 16. Experiments consisted of removing tumors from patient-derived hydrogel kidney phantoms. (**A**) Margin size is plotted in millimeters for each group. Signed margin error targeting a $6\,\mathrm{mm}$ margin is shown on the left axis for both Multi-shot and Single-shot (positive and negative values indicate larger and smaller than planned margin sizes, respectively). Multi-shot used two phantoms per patient model, other groups used one. (**B**) Absolute margin error targeting a $6\,\mathrm{mm}$ margin is plotted in millimeters. Surgeon is excluded since there was no target margin size. (**C**) Consistency of resection via coefficient of variance (COV), defined as the ratio of standard deviation and mean margin as a percentage (lower values are better), shown per patient model. (**D**) Images show Patient 1 before and after resection, and evaluation after CT imaging of specimen and conversion to a surface mesh. An example of resection failure is shown.

Surface randomization during training leads to small improvements, although the effect is modest. The average tumor localization error decreases from $2.35 \pm 0.85\,\mathrm{mm}$ to $2.13 \pm 1.04\,\mathrm{mm}$. Similarly, the Hausdorff 95% distance is reduced from $3.39 \pm 1.21\,\mathrm{mm}$ to $3.30 \pm 1.25\,\mathrm{mm}$. The IoU increases from $0.65 \pm 0.06$ to $0.66 \pm 0.08$. Across all metrics, the standard deviation increases slightly when surface randomization is applied. This indicates that while the method may improve average performance, it also introduces greater variability. Future work should explore cut randomization with varying levels of Perlin noise and with randomization of the CSI phase.

Our study does not quantify the sim-to-real accuracy of the physics-based tumor resection. While such analysis would have been insightful to isolating the effect of the simulation in the occupancy predictions, the required methodology is beyond the scope of this work. Nevertheless, the demonstrated occupancy network accuracy during the PDD phase provides evidence that our simulation captures accurate deformation in tumor resection.

### B. *Robotic Resection Performance Analysis*

The results demonstrate Multi-shot's ability to successfully resect hydrogel kidney tumors. Our intuition is that Single-shot fails every procedure due to a lack of understanding margin tissue stretching during vacuum gripper retraction. Without vision-based feedback and physics-based scene understanding, Single-shot underestimates remaining margins and cuts through the tumor. Multi-shot receives iterative feedback of the scene deformation enabling reduced margin error versus Single-shot. Adding a margin offset would have eventually allowed Single-shot to succeed without PSM, however, it requires predicting a patient-specific offset preoperatively.

Multi-shot and Surgeon both successfully resected all tumors with NSMs. In our experiments, Multi-shot removes more consistent margin sizes as indicated by the COV. On the other hand, the surgeon preserves more healthy kidney tissue. The surgeon's key challenge is intraoperatively estimating relative tumor boundary position during deep dissection stages.

In these stages the tumor boundary is occluded by healthy tissue, and lacking localization, the surgeon resects along a curve informed by the preoperative CT. This can be observed by Surgeon's margin sizes having the most variation of any group while also cutting within $1\,\mathrm{mm}$ of the tumor boundary in each procedure. For example, patient 4's more endophytic tumor made localization difficult and the surgeon chose to remove excessive margins to ensure success. Multi-shot's occupancy network continually updates scene understanding, providing guidance even during deep dissection stages.

Dexterity constraints of the robot made some tumor locations more difficult to access than others. For instance, patient 4's tumor placement led to awkward trajectory planning and resulted in surface incisions that deviate from the simulated training data. We believe that improving dexterity, either through improved robot positioning or by adding wristed degrees of freedom to the tools, could mitigate these issues and lead to better outcomes.

The $6\,\mathrm{mm}$ margin selection causes Multi-shot to remove more kidney tissue than Surgeon. This study's primary goal was resection success, future work will optimize the margin size to preserve more kidney tissue. While Multi-shot's vision pipeline is low-latency, repeated back-and-forth posture transitions between imaging and cutting extended procedure duration. Future work will investigate how to maintain vision guidance without switching between robotic arm postures. For example, by using a flicking motion of the end-effector which moves the depth sensor over the operational site temporarily, or by using an additional robotic arm.

Two types of events were logged during Multi-shot experiments: supervisor re-plan requests and vacuum-grasp failures. First, supervisor re-planning was occasionally triggered when the occupancy network's output substantially differed from the supervisor's perception of the real world, caused by artifacts in the observation point cloud resulting from background segmentation errors. Future work will focus on more robust segmentation methods; additionally, the uncertainty values provided by LUDO could be used to automatically trigger re-planning [36]. Second, three vacuum-grasp failures occurred with Patient 2 due to the small tumor apex reducing the bellows pad sealing area; each was resolved by manual re-engagement, after which the autonomous procedure continued to achieve an NSM. Future work will explore automated detection and recovery, such as vacuum pressure monitoring.

The Single-shot emergency halt was a supervisor-initiated Orocos RTT deployer termination triggered by unacceptable tool proximity between the electrosurgical tool tip and vacuum pad; no collision occurred. This contributed to reporting only three Single-shot trials despite four being planned.

### C. Limitations

Despite the success of this study, several technical and clinical limitations remain. First, the model-based progressive vacuum retraction trajectory may not generalize to other tumor resection procedures. Future work will incorporate online planning with vision-based feedback, incorporating learning-based tissue manipulation methods such as imitation learning [58] and deep reinforcement learning [59] for adaptive retraction. Second, the occupancy network cannot account for scene appearances unanticipated by the simulation. While simulation was feasible in this study due to the predictable state-based robot system, extending this approach to other resections will require new simulations and anticipating tissue deformations.

Third, we used a non-perfused hydrogel phantom that does not bleed intraoperatively. Future work should design perfused phantoms following Melnyk et al. [41], and improve FSM-based autonomy to handle unexpected events such as bleeding and perform hemostatic control.

Fourth, patient tumor size and shape may differ between preoperative imaging and surgery which the occupancy network currently cannot account for. Future work will explore multimodal guidance that fuses occupancy predictions with intraoperative US and NIRF, e.g., by encoding segmented US or NIRF cues as additional network inputs, using US-based elastography for subject-specific adaptation [60], and treating modality mismatch as a surgeon-facing redundancy check. Robotic palpation [61] may further help for endophytic tumors. Moreover, rather than using patient-specific occupancy networks requiring per-case retraining, future work should explore generalized or zero-shot resection tracking conditioned on preoperative data at inference.

Fifth, this robotic study is limited in its ability to statistically compare tumor resection methods. While comparing sampled margin points, summarized in Table IV, would appear to show statistical significance, these points are spatially correlated and therefore cannot be considered independent samples. When aggregating metrics across removed kidney tumors, summarized in Table V, the two Multi-shot resections per patient model cannot be treated as independent; treating the data as paired by patient model and averaging Multi-shot results reduces effective sample sizes to $N = 3$ for comparing with Single-shot (due to one aborted trial) and $N = 4$ for comparing with Surgeon. These sample sizes also cause non-parametric tests like Wilcoxon signed-rank tests on resection consistency (COV) to be underpowered. Per-cut analysis could leverage more samples, as shown in Table VI, Multi-shot performs 77 cuts across eight procedures. However, consecutive cuts within a procedure are inherently dependent as later cuts may reshape the geometry from prior cuts. Additionally, CT-based margin assessment requires the complete resected specimen, and not all PSM sites can be reliably mapped to specific cuts. Clinically, resection outcome analysis often focuses on the minimum margin, and related surgical studies rely on substantially larger sample sizes to ensure meaningful statistical analysis [62]. A larger study involving more tumor resections is the subject of future work.

### D. Future Clinical Translation

Future clinical translation can follow two largely separate pathways: extending the present AROPN system toward *in vivo* OPN validation, or pursuing autonomous robotic minimally invasive PN through substantial system redesign.

While our experiments evaluate tumor resection during open surgery on hydrogel phantoms, gaps remain before translation to real OPN. As discussed in Section IX-C, the present study does not reproduce blood perfusion or preoperative-to-intraoperative anatomical mismatch. Future work will therefore conduct *in vivo* porcine studies using transgenic tumor models such as Oncopig [38], which enables site-specific renal tumor induction with realistic biomechanical properties including anisotropy and viscoelasticity. While our current kidney phantom mimics fresh porcine kidney mechanical properties [41], *in vivo* properties vary across subjects; subject-specific adaptation may leverage the multimodal and elastography strategies outlined in Section IX-C.

Migrating from OPN to minimally invasive PN is a separate long-term direction with four challenges. First, the current RGB-D sensor must be replaced by laparoscopic depth estimation using SLAM/SfM [63] or a stereo foundation model [64]. Second, under minimally invasive PN, peritumoral fat, surrounding organs, tool occlusion, and electrosurgical smoke challenge segmentation and occupancy inference, requiring more robust vision pipelines. Third, straight tools must be replaced with wristed laparoscopic instruments and software-based remote-center-of-motion control to operate through fixed trocar ports with adequate dexterity. Fourth, the collision and planning layers must be extended, requiring the planner to distinguish tissue to be cut from surrounding healthy structures that become hard obstacles with additional safety margins.

## X. Conclusion

This work presents a supervised autonomous robotic workflow for margin-controlled soft-tissue tumor resection in patient-derived hydrogel kidney phantoms under an OPN setting.. Our first contribution introduces a Multi-shot occupancy-guided intraoperative inference workflow that employs a data-driven occupancy network to reliably guide the full course of resection. The network is trained using a physics-based tumor resection simulation derived from preoperative patient CT scans, enabling the system to localize the tumor, margin tissue, and kidney from single-view partial point clouds despite iterative tissue deformation and dissection. The inference workflow achieves a tumor centroid error of $2.13 \pm 1.04\,\mathrm{mm}$ compared to reference CT volumes across 16 states of tumor resection. Notably, this framework maintains effective tumor tracking even when the tumor is fully occluded by margin tissue during deep margin dissection periods as illustrated in Figure 3, with runtime summarized in Table VII, addressing a key limitation of existing intraoperative imaging methods such as US and NIRF. For our second contribution, we developed a hybrid learning and model-based planning framework for autonomy tailored to the unique constraint in tumor resection procedures: inadvertent cuts into the tumor that compromise oncologic outcomes often cannot be detected or corrected intraoperatively, making first-attempt success critical. This approach combines the adaptability of learning-based scene understanding with the reliability and oversight of a structured state machine model under human supervision. As our third contribution, the proposed inference and planning frameworks are integrated into a dual-arm robot system capable of electrosurgical cutting and vacuum-based tissue manipulation, and successfully achieved accurate autonomous tumor resections using patient-derived hydrogel phantoms.

## Appendix

We provide supplemental analysis of our vision pipeline.

### A. Timings

To quantify runtime of our vision pipeline, we report the latency of the perception pipeline for the CSI and PDD in Table VII. Note, inference times of occupancy networks are not affected by the shape of the object or scene that they represent. However, the query point count affects their inference time. The effect of the query point count on inference time is discussed in [36].

TABLE VII
Perception runtime for vision pipeline. Latencies are jointly reported for the CSI and PDD models. Hardware: Intel Core i9-14900KF (CA, USA), NVIDIA GeForce RTX 4080 (CA, USA)

| Stage | Time | Notes |
|---|---|---|
| Point cloud capture | $0.2629 \pm 0.0060\,\mathrm{s}$ | Depth sensor acquisition |
| SAM2 segmentation | $0.0833 \pm 0.0900\,\mathrm{s}$ | Kidney point extraction |
| Occupancy inference | $0.0072 \pm 0.0008\,\mathrm{s}$ | $10^6$ query points |

### B. Cut Surface Randomization

TABLE VIII
Optimal margin identification and tumor localization error during the PDD phase on ablation occupancy networks trained without cut surface randomization (see Figure 6). Margin identification evaluates the ability to identify the appropriate margin for resection as absolute and signed error from desired 6 mm. Centroid error, Hausdorff 95%, and IoU for tumor localization. N=16 samples.

| Patient | Optimal Margin Identification | | Tumor Localization | | |
|---|---|---|---|---|---|
| | Abs. Error (mm) | Sig. Error (mm) | Centroid Error (mm) | Hausdorff 95% (mm) | IoU |
| 1 | $2.39 \pm 1.91$ | $+1.79 \pm 2.48$ | $2.84 \pm 1.49$ | $3.69 \pm 1.75$ | $0.65 \pm 0.12$ |
| 2 | $1.30 \pm 0.81$ | $+1.17 \pm 0.99$ | $2.21 \pm 0.50$ | $2.40 \pm 0.33$ | $0.63 \pm 0.04$ |
| 3 | $2.70 \pm 1.88$ | $+2.40 \pm 2.25$ | $2.37 \pm 0.79$ | $4.49 \pm 0.79$ | $0.73 \pm 0.04$ |
| 4 | $1.72 \pm 1.08$ | $-1.51 \pm 1.56$ | $2.01 \pm 0.25$ | $2.97 \pm 0.51$ | $0.59 \pm 0.04$ |
| Average | $2.06 \pm 1.61$ | $0.34 \pm 2.66$ | $2.35 \pm 0.85$ | $3.39 \pm 1.21$ | $0.65 \pm 0.06$ |

### C. Segmentation Artifacts

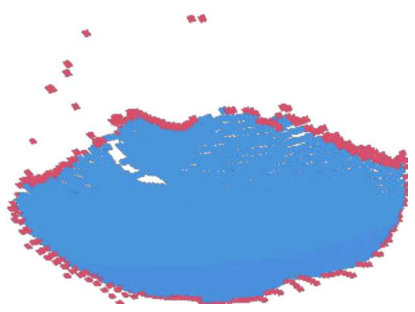

Fig. 17. Example of how point clouds are processed. The red points are removed by an erosion of the SAM 2 segmentation mask.

While this study does not focus on image segmentation of the kidney from the background, we seek to evaluate the effect that segmentation errors have on occupancy performance. Our segmentation pipeline applies a binary erosion to the SAM 2 segmentation mask to suppress boundary artifacts in depth images, where depth drop-off near object edges can introduce outlier points into the resulting point cloud. For this evaluation, we apply SAM 2 segmentation to images from the PDD states of the CT-Zivid dataset without the $5 \times 5$ kernel erosion, resulting in spurious boundary points being passed to the occupancy network, see Figure 17. We evaluate the centroid error of the occupancy predicted tumor relative to the CT tumor. Results show an average $13.16 \pm 11.29\,\mathrm{mm}$ centroid error, compared to results in Table III with average centroid error of $2.13 \pm 1.04\,\mathrm{mm}$. These results highlight the importance of robust segmentation and point cloud filtering for accurate tumor localization, particularly in the more challenging segmentation conditions encountered in real surgical settings.

### D. Observation Angle

In our real world experiments, the Zivid 2+ M60 Camera always observed the kidney scene from the same observation angle. However, different observation angles may provide differing information in the input point cloud for predicting the tumor location. To evaluate this, we use the four CT images obtained from the PDD 3-D anatomy inference evaluation (see Section VII-B) for patient 2 and collect simulated point cloud observations following our training data generation setup (see Figure 7) with varying observation angles (up to $\pm 35°$ for each axis). We collect $100$ simulated depth point cloud captures for all $4$ PDD resection states for a total of $400$ inferences. Results show the average centroid error of the tumor compared to the reference CT tumor as $1.92 \pm 0.54\,\mathrm{mm}$, compared to the single-angle real world centroid error from Table III of $2.14 \pm 0.36\,\mathrm{mm}$. These results demonstrate the robustness of the occupancy network to varying observation angles and suggest that an improved camera angle could have improved our evaluation. Methods to find such a camera angle should be explored in future work.

### E. Simulated Occupancy Evaluation

To evaluate the effect of sim-to-real transfer on occupancy network performance, we test the accuracy of the occupancy predictions on unseen simulation states from the physics-based tumor resection simulation. We create a test dataset according to Section VII-B with 400 new simulation states for patient 2. Note, this new simulation data was absent from the training data. We evaluate the centroid error of the occupancy predicted tumor to the reference tumor in simulation. Results give an average centroid error of $0.92 \pm 0.42\,\mathrm{mm}$, compared to the real world centroid error from Table III of $2.14 \pm 0.36\,\mathrm{mm}$. This shows an increase in centroid error of $1.22\,\mathrm{mm}$ when translating from the simulation environment to the real world.

### F. Observation Occlusion

We conducted experiments to evaluate the occupancy network performance under various occlusion scenarios.Four Zivid point cloud captures were taken of a hydrogel kidney phantom in a partially resected state under various occlusions by surgical instruments and fake blood. The results measure the standard deviation of the occupancy network's predicted tumor centroid about its mean: unoccluded scene $\pm 0.49\mathrm{mm}$, surgical instruments occlusion $\pm 1.85\mathrm{mm}$, blood occlusion $\pm 0.78\mathrm{mm}$, surgical instruments+blood occlusion $\pm 1.51\mathrm{mm}$.

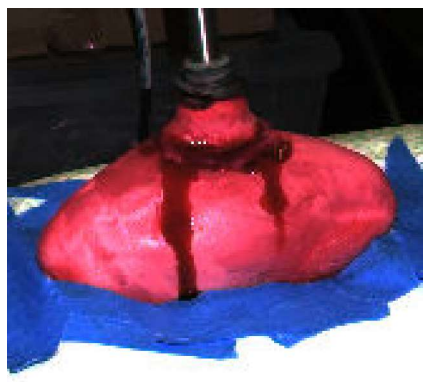

Bleeding

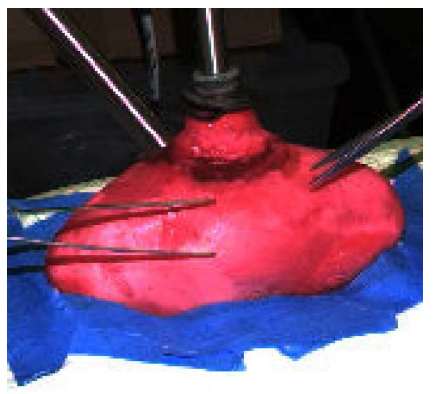

Surgical Instruments

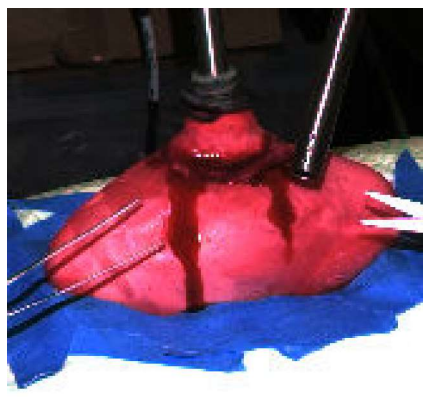

Bleeding+Instruments

Fig. 18. Example scenes for observation occlusion tests using a hydrogel kidney phantom, fake blood, and surgical instruments.